%% file: main-nv.tex
\pdfoutput=1
\documentclass[10pt,logo,copyright]{nvidiatechreport}
\input{packages}

\definecolor{darkred}{rgb}{0.7,0.0,0.0}

\usepackage{pifont}
\usepackage{wrapfig}

\usepackage[nameinlink]{cleveref}
\crefname{equation}{Eq.}{Eqs.}
\crefname{figure}{Fig.}{Figs.}
\crefname{section}{Sec.}{Sec.}
\crefname{appendix}{App.}{App.}
\crefname{table}{Tab.}{Tabs.}
\crefname{algorithm}{Algo}{Algo}
\crefname{thm}{Thm}{Thm}
\Crefname{thm}{Thm}{Thm}
\crefname{prop}{Prop}{Prop}
\usepackage{longtable}
\usepackage{mathtools}
\usepackage{siunitx}
\usepackage{booktabs}
\usepackage[table,xcdraw]{xcolor}

\usepackage{booktabs,tabularx,pifont,array,bbm,float,makecell,multirow}
\usepackage{ragged2e}
\definecolor{nvidiagreen}{HTML}{76B900}
\definecolor{bestrow}{HTML}{E1EBD7}
\newcolumntype{Y}{>{\raggedleft\arraybackslash}X}
\newcolumntype{L}[1]{>{\raggedright\arraybackslash}p{#1}}

\input{macros}

\newcommand{\crefnames}[3]{%
  \@for\next:=#1\do{%
    \expandafter\crefname\expandafter{\next}{#2}{#3}%
  }%
}

\title{\vspace{5mm} \name: Unifying Parametric Human Body Models}

\author{

 Jun Saito$^{\ast}$, Jiefeng Li$^{\ast}$, Michael de Ruyter$^{\ast}$, Miguel Guerrero, Edy Lim, Ehsan Hassani, Roger Blanco Ribera, Hyejin Moon, Magdalena Dadela, Marco Di Lucca, Qiao Wang, Xueting Li, Jan Kautz, Simon Yuen, Umar Iqbal$^{\ast}$\\
  \small NVIDIA \\
  \small $^{\ast}$Core Contributors \quad \\
  {\small \href{https://github.com/NVlabs/SOMA-X}{\texttt{https://github.com/NVlabs/SOMA-X}}}
}

\makeatother

\input{sections/00_abstract}

\begin{document}
\maketitle
\vspace{-8mm}
\begin{figure}[ht!]
    \centering
    \includegraphics[trim={0.45cm 0.36cm 0.4cm 0},clip,width=1.05\textwidth]{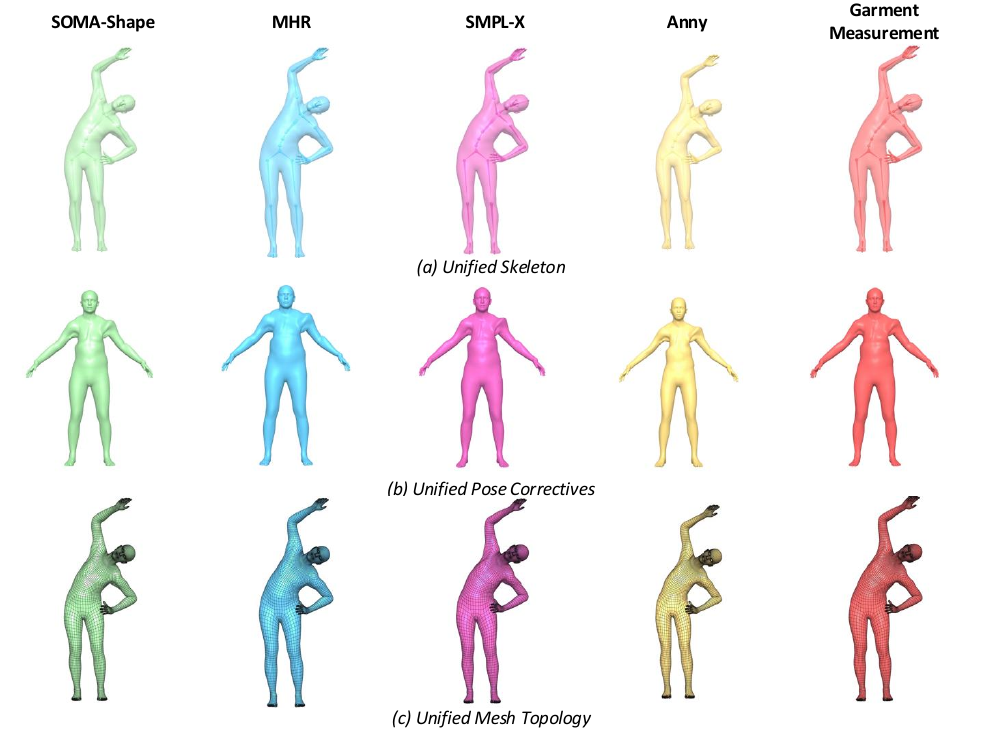}
    \caption{\name unifies five heterogeneous parametric body models (SOMA-Shape, MHR, SMPL-X, Anny, and GarmentMeasurements) under a single animation pipeline.
    \textbf{(a) Unified Skeleton:} despite originating from entirely different identity spaces, joint hierarchies, and mesh resolutions, all five models are driven by the same \name{} skeleton in an identical pose, with no model-specific retargeting.
    \textbf{(b) Unified Pose Correctives:} a single MLP correctives model trained once on the shared \name{} topology produces anatomically plausible pose-dependent deformations for all backends, mitigating standard LBS artifacts without per-model corrective learning.
    \textbf{(c) Unified Mesh Topology:} all identity models share the same mesh structure, enabling skinning weights, deformation priors, and correctives to transfer seamlessly across backends.}
    \label{fig:teaser}
\end{figure} 

\vspace{-9mm}

\blfootnote{\scriptsize
  \noindent\textbf{Related projects that already support SOMA:}
  \begin{itemize}
    \setlength{\itemsep}{0pt}    
    \setlength{\parskip}{0pt}    
    \item \textbf{SOMA Retargeter} for SOMA to humanoid retargeting:    {\small \href{https://github.com/NVIDIA/soma-retargeter}{\texttt{https://github.com/NVIDIA/soma-retargeter}}}
    \item \textbf{GEM} for video pose estimation:  {\small \href{https://github.com/NVlabs/gem-x}{\texttt{https://github.com/NVlabs/gem-x}}}
    \item \textbf{Kimodo} for controllable text-to-motion generation:   {\small \href{https://github.com/nv-tlabs/kimodo}{\texttt{https://github.com/nv-tlabs/kimodo}}}    
    \item \textbf{BONES-SEED} is the largest human(oid) motion dataset (150k motions):  
    {\small \href{https://huggingface.co/datasets/bones-studio/seed}{\texttt{https://huggingface.co/datasets/bones-studio/seed}}}
    \item \textbf{ProtoMotions} is a simulation and learning framework for human(oids):  
    {\small \href{https://github.com/NVlabs/ProtoMotions}{\texttt{https://github.com/NVlabs/ProtoMotions }}}
    \item \textbf{GEAR SONIC} is a humanoid behavior foundation model:  
    {\small \href{https://github.com/NVlabs/GR00T-WholeBodyControl}{\texttt{https://github.com/NVlabs/GR00T-WholeBodyControl}}}
  \end{itemize}
}

\clearpage
\abscontent

\input{sections/01_introduction}
\input{sections/02_related_work}

\input{sections/03_method}
\input{sections/04_experiments}

\input{sections/05_conclusion}


\section*{Acknowledgments}
We thank Davis Rempe, Mathis Petrovich, and Sehwi Park for valuable feedback and help throughout the project.
We thank Cyrus Hogg and Mike Sandrik for their support with data acquisition, and Dennis Lynch, Will Tellford, Jon Shepard, and Spencer Huang for helpful guidance during development.


\appendix
\newpage
\clearpage

\clearpage 

\setcitestyle{numbers}
\bibliographystyle{plainnat}
\bibliography{main.bib}

\end{document}

%% file: packages.tex
\usepackage[authoryear,sort&compress,round]{natbib}

\usepackage[utf8]{inputenc} 
\usepackage[T1]{fontenc}    

\usepackage{parskip}        
\usepackage{url}            
\usepackage{hyperref}       
\usepackage{booktabs}       
\usepackage{amsfonts}       
\usepackage{nicefrac}       
\usepackage{microtype}      
\usepackage{xcolor}         
\usepackage[dvipsnames]{xcolor} 
\usepackage{graphicx}
\usepackage{animate}        
\usepackage{subcaption}
\usepackage{tabularx}
\usepackage{makecell}
\usepackage{adjustbox}
\usepackage{setspace}
\newcolumntype{M}[1]{>{\centering\arraybackslash}m{#1}}
\usepackage{float}
\usepackage{tikz}
\usetikzlibrary{positioning,shapes,arrows}
\usepackage{amsmath,amsfonts,bm, bbm,leftindex}
\usepackage{multirow}
\usepackage{comment}
\usepackage{gensymb}
\usepackage{lipsum}
\usetikzlibrary{arrows.meta, positioning, fit}
\usepackage[para]{threeparttable}
\usepackage{tikz}
\usetikzlibrary{tikzmark}



%% file: macros.tex
\usepackage{amsfonts}
\usepackage{bm}
\usepackage{amsmath}
\usepackage{mathtools}
\usepackage{multirow}
\usepackage{booktabs}
\usepackage{caption}
\usepackage{enumitem}
\usepackage{wrapfig,lipsum,booktabs}
\usepackage{caption}
\usepackage{bbm}
\usepackage{multirow}
\usepackage{pifont}
\usepackage[linesnumbered,ruled,vlined]{algorithm2e}

\SetCommentSty{mycommfont}
\SetAlCapFnt{\small} 
\usepackage{etoc}
\SetAlgorithmName{Algo}{Algo}{}

\renewcommand{\paragraph}[1]{{\vspace{1mm}\noindent \bf #1}.}

\DeclareMathOperator*{\argmin}{arg\,min}

\newcommand{\name}{\text{SOMA}\xspace}

\newcommand\blfootnote[1]{%
  \begingroup
  \renewcommand\thefootnote{}\footnote{#1}%
  \addtocounter{footnote}{-1}%
  \endgroup
}

%% file: sections/00_abstract.tex
\begin{abstract}
Parametric human body models are foundational to human reconstruction, animation, and simulation, yet they remain mutually incompatible: SMPL, SMPL-X, MHR, Anny, and related models each diverge in mesh topology, skeletal structure, shape parameterization, and unit convention, making it impractical to exploit their complementary strengths within a single pipeline.
We present \name, a unified body layer that bridges these heterogeneous representations through three abstraction layers.
\textit{Mesh topology abstraction} maps any source model's identity to a shared canonical mesh in constant time per vertex.
\textit{Skeletal abstraction} recovers a full set of identity-adapted joint transforms from any body shape, whether in rest pose or an arbitrary posed configuration, in a single closed-form pass, with no iterative optimization or per-model training.
\textit{Pose abstraction} inverts the skinning pipeline to recover unified skeleton rotations directly from posed vertices of any supported model, enabling heterogeneous motion datasets to be consumed without custom retargeting.
Together, these layers reduce the $O(M^2)$ per-pair adapter problem to $O(M)$ single-backend connectors, letting practitioners freely mix identity sources and pose data at inference time.
The entire pipeline is fully differentiable end-to-end and GPU-accelerated via NVIDIA-Warp.
\keywords{Parametric Human Body Model \and Skeleton Fitting \and Mesh Topology Abstraction \and Pose Inversion \and Identity-Pose Decoupling \and Linear Blend Skinning}

\end{abstract}

%% file: sections/01_introduction.tex
\section{Introduction}
\label{sec:intro}

Parametric human body models are a cornerstone of computer vision, computer graphics, and physical AI, enabling reconstruction, animation, and simulation of human motion at scale.
The most widely used family, SMPL~\citep{SMPL:2015,SMPL-X:2019},  defines meshes with compact linear shape spaces and has become the de facto target for pose estimation, motion generation, and avatar synthesis~\citep{shen2024gvhmr, wang2026duomo, zhang2024large, li2025genmo}. 
MHR~\citep{MHR:2025} introduces explicit bone-length parameterization that more faithfully captures skeletal diversity across people, addressing a well-known limitation of PCA-only shape models.
Anny~\citep{bregier2025anny} constructs its shape space from anthropometric measurements rather than 3D scans, providing semantic control over age, height, weight, and body composition that spans the full human lifespan from infants to elders, aiming to remove the demographic biases inherent in scan-collected data.
GarmentMeasurements~\citep{GarmentCode2023} extends shape representation to clothing-aware body proportions encoded via body measurements.

Despite this diversity of available models, a concrete fragmentation problem persists.
Each model defines its own mesh topology, joint hierarchy, unit convention, and parameter space.
A practitioner wishing to combine Anny's interpretable phenotype control with an SMPL-compatible motion capture dataset must implement separate topology-transfer pipelines, independent skeleton-fitting routines, and bespoke coordinate-frame conversions for every model pair.
Supporting $M$ models naively requires $O(M^2)$ per-pair adapters; in practice, this forces early model commitment and forfeits the complementary strengths of alternatives.
No unified interface currently exists that lets a researcher freely mix identity source and pose parameterization.

We introduce \name{}, a canonical body topology and rig that serves as a universal pivot for heterogeneous parametric body models (see Fig.~\ref{fig:overview}).
Rather than replacing existing models, \name{} maps their rest-shape outputs to a single shared representation, after which any identity model can be animated through one unified LBS pipeline.
This reduces the $O(M^2)$ adapter problem to $O(M)$ single-backend connectors, each implemented once and composed freely at inference time.

\noindent Our contributions are fourfold:

\begin{enumerate}
    \item \textbf{Identity-Pose Decoupling via a Canonical Topology (\texttt{SOMALayer}).}
    We propose a framework that maps the rest-shape output of any supported parametric model to a canonical \name{} mesh and rig, explicitly separating identity representation from kinematic parameterization.
    A single pose interface drives any identity source without model-specific adaptation code at runtime, and a unified pose-dependent correctives model generalizes to all backends.

    \item \textbf{Mesh Topology Abstraction.}
    We introduce a topology abstraction module that pre-computes a fixed 3D barycentric correspondence between each source model's neutral mesh and the \name{} canonical mesh at initialization.
    At runtime, identity transfer requires no neural forward pass and no iterative solver.

    \item \textbf{Skeletal Abstraction.}
    We present a backend-agnostic skeleton fitting algorithm that exploits the shared mesh correspondence to precisely fit any template skeleton into a new body shape.
    Given the transferred rest shape, it recovers identity-adapted world-space joint transforms in a single analytical forward pass with no iterative optimization or per-model training.

    \item \textbf{Pose Abstraction.}
    We introduce a pose abstraction module that recovers \name{} skeleton rotations from posed vertices of any supported model via analytical inverse-LBS with Newton-Schulz orthogonalization, enabling direct conversion of motion data from SMPL, MHR, and other models into \name{}'s unified skeleton convention.
\end{enumerate}
The entire \name{} forward pass is fully differentiable end-to-end, making it directly usable as a differentiable layer in large-scale optimization and ML training pipelines.

%% file: sections/02_related_work.tex
\section{Related Work}
\label{sec:related}

\subsection{Parametric Body Models}

The field has seen significant evolution in parametric human modeling~\citep{anguelov2005scape, SMPL:2015, pishchulin17pr, osman2020star, ghum, MHR:2025, bregier2025anny}.
\textbf{SMPL}~\citep{SMPL:2015} introduced a vertex-based linear blend skinning model with learned corrective blend shapes, which became the de facto standard with 6{,}890 mesh vertices and a compact PCA shape space.
\textbf{STAR}~\citep{osman2020star} proposed sparser skinning weights to reduce undesirable cross-joint coupling.
SMPL-H~\citep{MANO:SIGGRAPHASIA:2017} and \textbf{SMPL-X}~\citep{SMPL-X:2019} extended the SMPL family with fully articulated hands, via \textbf{MANO}~\citep{MANO:SIGGRAPHASIA:2017}, and an expressive face, respectively. 
\textbf{MHR}~\citep{MHR:2025} addresses skeletal ambiguity by explicitly modeling bone lengths to improve fitting accuracy across body proportions.
\textbf{Anny}~\citep{bregier2025anny} builds its shape space from anthropometric measurements rather than 3D scans, enabling phenotype control (age, height, weight) that spans infants to elders.
Each of these models defines its own mesh topology, joint hierarchy, and parameter space.
\name{} does not replace any of them; instead, it provides a canonical mesh topology and rig that any supported backend can drive through a single unified pipeline.

\subsection{Human Motion Estimation and Generation}

A rich body of work estimates 3D human pose and shape from monocular images~\citep{hmr, iqbal2021kama, spin, pare, hmr2, yuan2022glamr, camerahmr, kocabas2024pace, nlf, wang2025prompthmr}, videos~\citep{vibe, tcmr, goel2023humans, wham, shen2024gvhmr, wang2026duomo}, and generates motion from diverse conditioning signals such as text, music, and scene context~\citep{tevet2023human, yuan2023physdiff, zhang2022motiondiffuse, zhang2024large, petrovich24stmc, li2025genmo}.
The vast majority of these systems are built around SMPL or SMPL-X as the target representation; more recently, methods such as MultiHMR~\citep{multihmr}, Sam-3D-Body~\citep{yang2026sam3dbody}, and DuoMo~\citep{wang2026duomo} have started to adopt MHR and Anny to better capture bone-length and age-range diversity.
However, whether estimating or generating, all of these systems are trained to output parameters for one specific body model and must be retrained whenever the target representation changes.
\name{} decouples identity model selection from the estimation pipeline: a pose estimator or generative model outputting \name{}-compatible joint parameters can drive any supported identity, SMPL, MHR, Anny, or others, at inference time without retraining, and can be supervised with body shape labels from all backends simultaneously. 
 


%% file: sections/03_method.tex
\section{Method}
\label{sec:method}

\begin{figure}[t]
\centering
\includegraphics[trim={0.3cm 0cm 0cm 0cm}, clip]{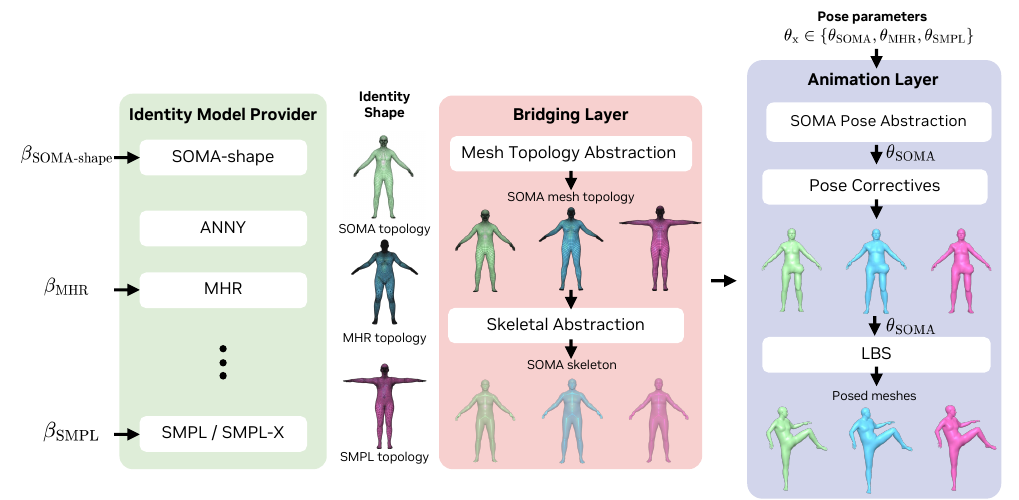}
\caption{%
\textbf{Overview of \name{}.}
\name{} decouples body identity from pose through three sequential layers.
\textit{Identity Model Provider} (left): any supported backend (SOMA-shape,
Anny, MHR, SMPL/SMPL-X, or GarmentMeasurements) maps its own shape parameters
$\beta_s$ to a rest-shape mesh in its native topology.
\textit{Bridging Layer} (middle): two abstraction steps canonicalize the
source identity into a unified representation.
Mesh Topology Abstraction transfers the rest shape to the shared \name{} topology
via pre-computed barycentric coordinates; Skeletal Abstraction then fits the
shared $J{=}77$-joint \name{} rig to the transferred rest shape in a single
closed-form pass, with no iterative optimization or per-identity training.
\textit{Animation Layer} (right): all identity models are animated through the
shared \name{} skeleton using $\theta_{\text{SOMA}}$ joint rotations.
When motion data arrives in another model's convention, \ie $\theta_\mathrm{x} \in \{\theta_{\text{MHR}}, \theta_{\text{SMPL}}, \ldots\}$,
Pose Abstraction converts it to $\theta_{\text{SOMA}}$ by analytically inverting
the LBS pipeline; this step is bypassed when pose is already in the \name{} convention.
A shared MLP Pose Correctives model then predicts pose-dependent vertex
displacements to correct LBS artifacts, and Linear Blend Skinning produces the
final posed mesh.
The entire pipeline is fully differentiable end-to-end.%
}
\label{fig:overview}
\end{figure}

\name{} is a modular framework for unified parametric body modeling.
Its core runtime component, \texttt{\texttt{SOMALayer}}, accepts shape parameters from any supported identity backend alongside pose parameters, and produces posed mesh vertices and joint positions in meters.
Fig.~\ref{fig:overview} illustrates the full pipeline.

\subsection{Overview and Notation}

Let $V_h \in \mathbb{R}^{N_h \times 3}$ denote the \name{} canonical mesh with $N_h$ vertices, $F_h$ its triangle faces, and $J = 77$ its joint count (excluding the root).
For each supported backend $s \in \{\text{NOVA, MHR, Anny, SMPL, SMPL-X, Garment}\}$, let $\mathcal{M}_s(\beta_s)$ denote the backend's rest-shape generator, which maps identity parameters $\beta_s$ to a source mesh $V_s \in \mathbb{R}^{N_s \times 3}$ in the backend's native unit.
\name's forward pass transforms any $(V_s, \theta)$ pair into posed \name{} mesh vertices via three sequential steps: (1)~mesh topology abstraction;  (2)~closed-form skeleton fitting; and (3)~LBS posing.
Every step is fully differentiable, so the entire pipeline can serve as a differentiable layer in optimization and learning frameworks.

\begin{figure}[t]
\centering
\begin{tabular}{c}
\includegraphics[trim={31cm 0cm 30cm 0cm}, clip, width=\linewidth]{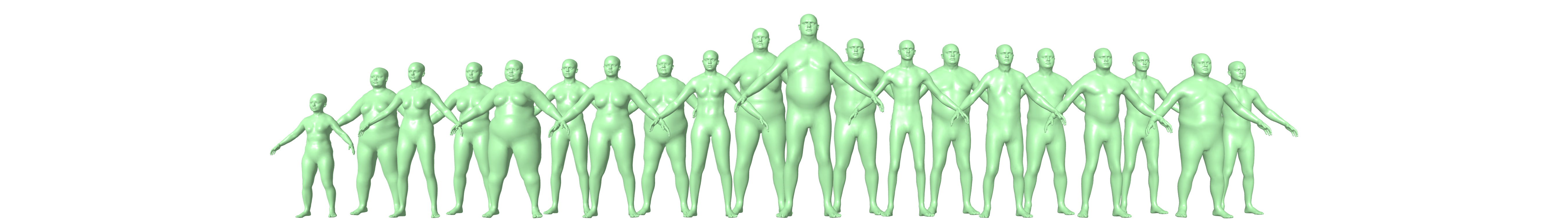} \\ 
(a) SizeUSA Registrations \\
\includegraphics[trim={25cm 0cm 24cm 0cm}, clip, width=\linewidth]{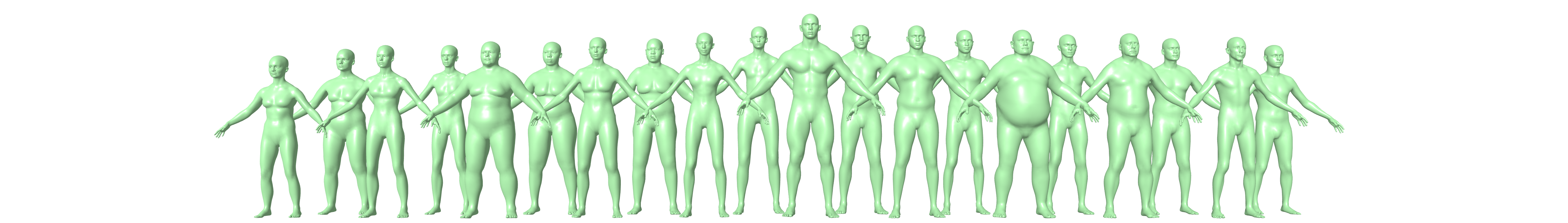} \\ 
(b) Triplegangers Registrations \\ 
\end{tabular}
\caption{%
\textbf{Training data for the \name{}-Shape PCA model.}
(a)~A subset of the 9,326 SizeUSA body scans registered to the \name{} topology, exhibiting a wide range of body weights and proportions.
(b)~A subset of the 303 Triplegangers scans, registered to the same topology.
All meshes are reposed to a canonical A-pose before PCA fitting, and mirror-augmented across the sagittal plane to enforce bilateral symmetry.%
}
\label{fig:sizeusa_registrations}
\end{figure}

\subsection{Identity Model Provider}
\label{sec:backends}


The identity model provider takes the native shape parameters $\beta_s$ of any supported backend and maps them to a rest-shape mesh in that backend's native topology.
\name{} integrates five interchangeable backends, each with its own strengths, allowing users to easily adopt the identity model of their preference within a single animation framework.

\paragraph{\name{}-Shape}
\name's own shape space uses $K{=}128$ PCA components trained on
9{,}326 SizeUSA body scans~\citep{SizeUSA2004}, 303
Triplegangers photogrammetry scans~\citep{Triplegangers}, and
samples distilled from the GarmentMeasurements PCA
model~\citep{GarmentCode2023}, with a 40/40/20 mixing ratio learned by incremental PCA~\citep{ross2008incremental}.  We show example registrations in Fig.~\ref{fig:sizeusa_registrations}.

\noindent\textbf{SMPL / SMPL-X}~\citep{SMPL:2015,SMPL-X:2019} parameterize body shape via PCA blend shapes (10 components for SMPL, 300 for SMPL-X) learned from registered 3D body scans and dominates research adoption. 

\noindent\textbf{MHR}~\citep{MHR:2025} parameterizes body shape via a combination of PCA identity coefficients and explicit bone-length scale factors that directly modulate skeletal proportions, providing fine-grained control over  body surface shape and skeletal proportions. 

\noindent{\textbf{Anny}}~\citep{bregier2025anny} parameterizes body shape using six anthropometric phenotypes (gender, age, muscle, weight, height, proportions) that drive multi-linear blendshapes spanning infants to elders. Anny is the sole backend capable of representing subjects below 18 years of age, hence making it the natural choice for applications requiring age-diverse or child-inclusive digital humans. 

\noindent\textbf{GarmentMeasurements}~\citep{GarmentCode2023} encodes body shape via 15 PCA components fitted to body scan data, similar to SMPL/SMPL-X, and is often adopted for garment modeling literature. 

\begin{figure}[t]
\centering
\includegraphics[trim={0cm 4.8cm 0cm 0cm},clip]{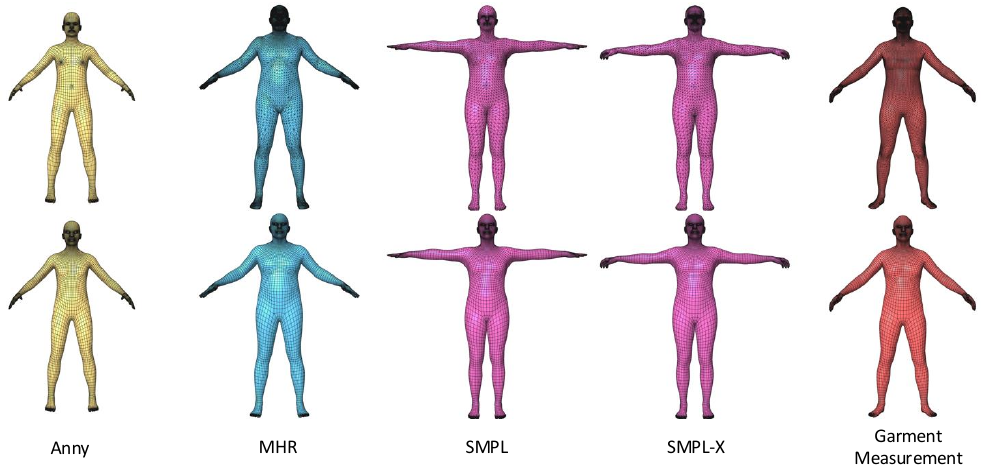}
\caption{%
\textbf{Mesh topology abstraction.}
Top: native mesh topologies of each identity model.
Bottom: the same identities mapped to the shared \name{} topology via 3D barycentric interpolation.
This common mesh serves as the pivot for all cross-model operations---skeleton fitting, pose transfer, correctives, and shape-space comparison all operate on a single canonical topology regardless of the source model.%
}
\label{fig:bary}
\end{figure}

\subsection{Mesh Topology Abstraction}
\label{sec:bary}

The mesh topology abstraction layer maps diverse source topologies from the identity model provider to the \name{} canonical mesh. We pre-compute a fixed 3D barycentric correspondence at initialization and apply it as a lightweight gather at runtime, as illustrated in 
Fig.~\ref{fig:bary} illustrates.

More specifically, given the source neutral mesh $(V_s, F_s)$ and a \name{} wrap mesh $V_h^{(s)}$ (a version of the \name{} template manually registered to the source model's neutral pose by an artist), we compute for each \name{} vertex $\mathbf{v}_i^h$ its 3D barycentric coordinates within a local tetrahedron of the source mesh.
For the closest source triangle $f_j = (u_1, u_2, u_3)$ to $\mathbf{v}_i^h$, we lift it to a tetrahedron by adding a fourth vertex along the face normal, $u_4 = u_1 + (u_2 - u_1) \times (u_3 - u_1)$, and solve for the barycentric coordinates $\mathbf{b} \in \mathbb{R}^4$ via a $3{\times}3$ linear system.
This tetrahedral lifting handles query points slightly off the surface without degeneracy.
Unlike 2D barycentric projection, 3D tetrahedral interpolation preserves volume in regions without clear surface correspondence, for example, when mapping between models with and without individual toes.
The pre-computation runs once at initialization; its output-, a face-index array $\mathbf{f} \in \mathbb{Z}^{N_h}$ and coordinate array $\mathbf{B} \in \mathbb{R}^{N_h \times 4}$, is stored as a fixed buffer.

At runtime, given deformed source vertices $V_s(\beta) \in \mathbb{R}^{N_s \times 3}$, each \name{} vertex is reconstructed as a weighted combination of its corresponding tetrahedron's vertices:
\begin{equation}
    \mathbf{v}_i^h(\beta) = \sum_{k=0}^{3} B_{ik} \cdot \tilde{V}_s(\beta)[\,\mathbf{F}^{\text{tet}}_{\mathbf{f}_i,\, k}\,],
\label{eq:bary_interp}
\end{equation}
where $\tilde{V}_s(\beta)$ is the source mesh augmented with one normal-offset point per face.
The cost is independent of the source vertex count $N_s$.

\begin{figure}[t]
\centering
\includegraphics[width=\textwidth]{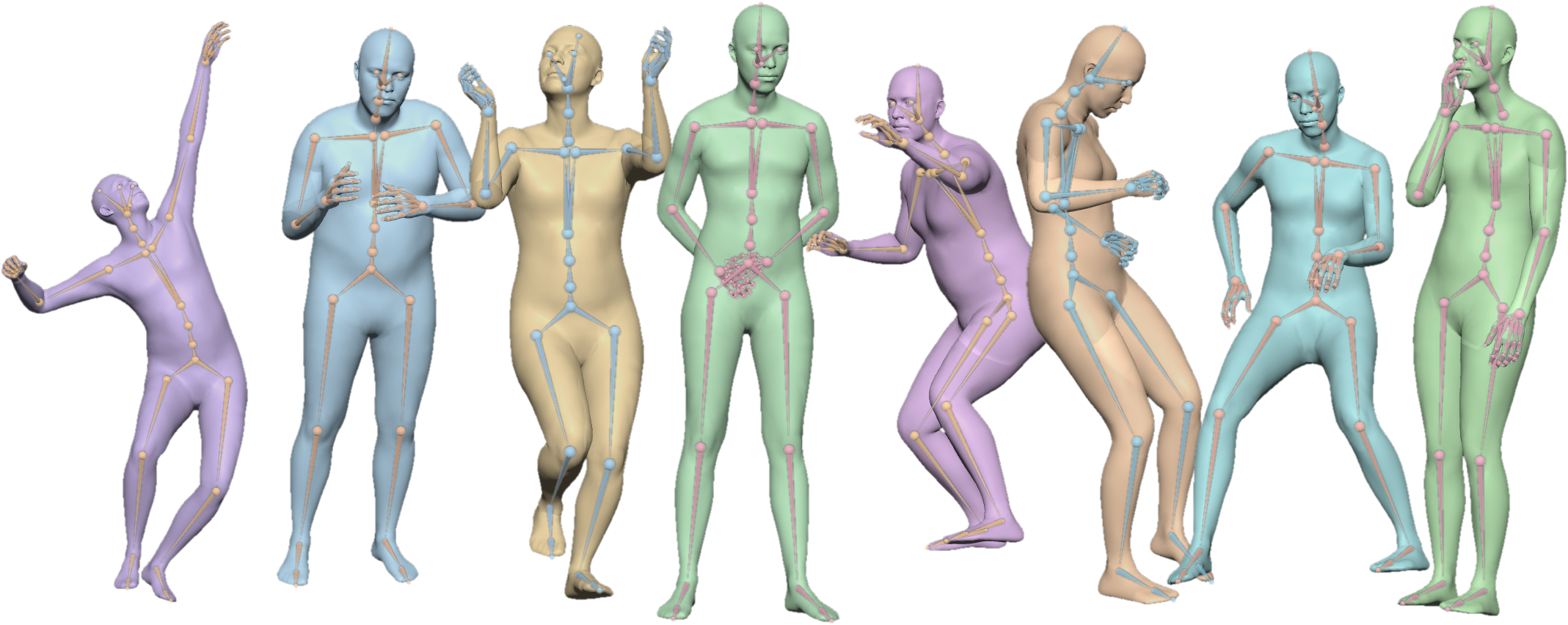}%
\caption{%
\textbf{Skeleton fitting on posed SAM~3D Body identities.}
Eight MHR identities in diverse poses with the \name{} skeleton fitted via \texttt{SkeletonTransfer} (\cref{sec:skeleton}).
Unlike joint regressors that assume a rest pose, our method generalizes to arbitrary posed shapes: joint positions are regressed via RBF interpolation, and joint rotations are recovered by Procrustes alignment, both in a single analytical forward pass with no iterative optimization.%
}
\label{fig:skelfitting}
\end{figure}

\subsection{Skeletal Abstraction}
\label{sec:skeleton}

Once all source meshes share a common topology (\cref{sec:bary}), we need a single skeletal structure to drive their pose.
We introduce \texttt{SkeletonTransfer}, a backend-agnostic algorithm that precisely fits any template skeleton into a new body shape given only the shared mesh correspondence.
In \name{}, we apply it to fit a $J{=}77$ joint rig; given a rest shape $V_h(\beta) \in \mathbb{R}^{N_h \times 3}$ on the \name{} topology, it recovers the full set of world-space joint transforms $\{T_k\}_{k=1}^J \subset SE(3)$ in two analytical stages: joint position regression and joint rotation fitting.
Fig.~\ref{fig:skelfitting} illustrates how the fitted skeleton adapts to bodies of varying proportions.

\subsubsection{Stage 1: Joint position regression via RBF}
For each joint $k \in \{1,\ldots,J\}$, we pre-build a per-joint Radial Basis Function regressor from the canonical bind-pose mesh.
The regressor uses the subset of vertices $\mathcal{N}_k$ that have non-zero skinning weight for joint $k$ or its parent.
Given the canonical bind-shape vertex positions $V^{\text{bind}}$, the RBF basis weights $\mathbf{w}_k$ are solved once by the linear system:
\begin{equation}
    \Phi(\mathcal{N}_k) \, \mathbf{w}_k = \mathbf{j}_k^{\text{bind}},
\label{eq:rbf_solve}
\end{equation}
where $\Phi$ is the linear RBF kernel evaluated at the neighborhood vertex positions, and $\mathbf{j}_k^{\text{bind}}$ is the canonical joint position.

At runtime, given identity rest shape $V_h(\beta)$, joint $k$'s world-space position is predicted as:
\begin{equation}
    \mathbf{j}_k(\beta) = \Phi\bigl(V_h(\beta)_{\mathcal{N}_k}\bigr) \, \mathbf{w}_k,
\label{eq:rbf_predict}
\end{equation}
which is a single linear operation.
All joint positions are computed in parallel via a pre-assembled sparse matrix $\mathbf{W}_{\text{RBF}} \in \mathbb{R}^{J \times N_h}$, reducing the full joint position update to one sparse matrix multiplication:
\begin{equation}
    J(\beta) = \mathbf{W}_{\text{RBF}} \, V_h(\beta)^T.
\label{eq:rbf_sparse}
\end{equation}

\subsubsection{Stage 2: Joint rotation fitting via Kabsch alignment}
Joint positions alone do not fully define the skeleton since each joint also requires an orientation that establishes its local coordinate frame.
Since source models assume different canonical poses (\eg,\ T-pose vs.\ A-pose), these orientations cannot be copied from the bind pose and must be fitted to the identity's rest shape.
With identity-adapted joint positions $\{{\mathbf{j}_k(\beta)}\}$ in hand, we recover the rotation component of each world-space joint transform via a two-step Kabsch alignment procedure.

\textit{Stage 2a: Inverse LBS initialization.}
For joint $k$, let $\mathcal{V}_k$ be the set of vertices with non-negligible skinning weight for joint $k$.
We estimate an initial global rotation $R_k^{\text{init}} \in SO(3)$ by solving the weighted orthogonal Procrustes problem~\citep{kabsch1976solution}:
\begin{equation}
    R_k^{\text{init}} = \argmin_{R \in SO(3)} \sum_{\mathbf{v} \in \mathcal{V}_k} \bigl\| R\bigl(\mathbf{v}^{\text{bind}} - \mathbf{j}_k^{\text{bind}}\bigr) - \bigl(\mathbf{v}(\beta) - \mathbf{j}_k(\beta)\bigr) \bigr\|^2.
\label{eq:procrustes}
\end{equation}

This is solved via SVD of the cross-covariance matrix.

\textit{Stage 2b: Child bone alignment.}
The initial rotation $R_k^{\text{init}}$ aligns the skinned vertex cloud but
may not correctly orient the bone vectors toward child joints.
We compute a refinement rotation $R_k^{\text{align}}$ that aligns the rotated
bind bone vectors $R_k^{\text{init}}(\mathbf{j}_c^{\text{bind}} - \mathbf{j}_k^{\text{bind}})$
to the target bone vectors $\mathbf{j}_c(\beta) - \mathbf{j}_k(\beta)$.
For joints with a single child (the majority of the skeleton), this is the
shortest-arc (Rodrigues) rotation between two vectors; for joints with multiple
children, we solve the Procrustes problem (\cref{eq:procrustes}) over the set
of child bone vectors.
The final world-space rotation is
$R_k = R_k^{\text{align}} \cdot R_k^{\text{init}} \cdot R_k^{\text{bind}}$,
where $R_k^{\text{bind}}$ is the canonical bind-pose world rotation,
and the complete transform is $T_k = \operatorname{SE3}(R_k, \mathbf{j}_k(\beta))$.
Both paths are fully vectorized across all joints via NVIDIA Warp custom
kernels, enabling massively parallel GPU execution with no sequential joint loop.

\subsection{Animation Layer}
\label{sec:lbs}

Given the identity-adapted joint transforms $\{T_k(\beta)\}$ and rest shape $V_h(\beta)$, \name's animation layer applies standard Linear Blend Skinning (LBS) and correctives to produce animated vertices.


\subsubsection{Posing}
Given input pose parameters (axis-angle $(B, 77, 3)$ or rotation matrices $(B, 77, 3, 3)$), together with an optional root translation $t_0 \in \mathbb{R}^3$, forward kinematics computes global joint transforms $\{G_k(\theta)\}_{k=0}^{J-1}$ by composing local rotations up the joint hierarchy.
\name{} can optionally apply \textit{joint orient} (pose-relative parameterization) where input rotations are expressed relative to the joint's canonical pose (usually $T$-pose or $A$-pose), which matches the convention of many parametric human models and their associated datasets.
Posed vertex positions are then:
\begin{equation}
    \mathbf{v}'_i = \sum_{k=0}^{J-1} w_{ik} \, G_k(\theta) \, T_k^{\text{bind}\,-1} \, \tilde{\mathbf{v}}_i,
\label{eq:lbs}
\end{equation}
where $\tilde{\mathbf{v}}_i$ is the homogeneous rest-shape position and $w_{ik}$ is the skinning weight.


\begin{figure*}[t!]
\centering
\begin{tabular}{c}
\includegraphics[trim={6cm 8cm 0cm 8cm}, clip,width=\textwidth]{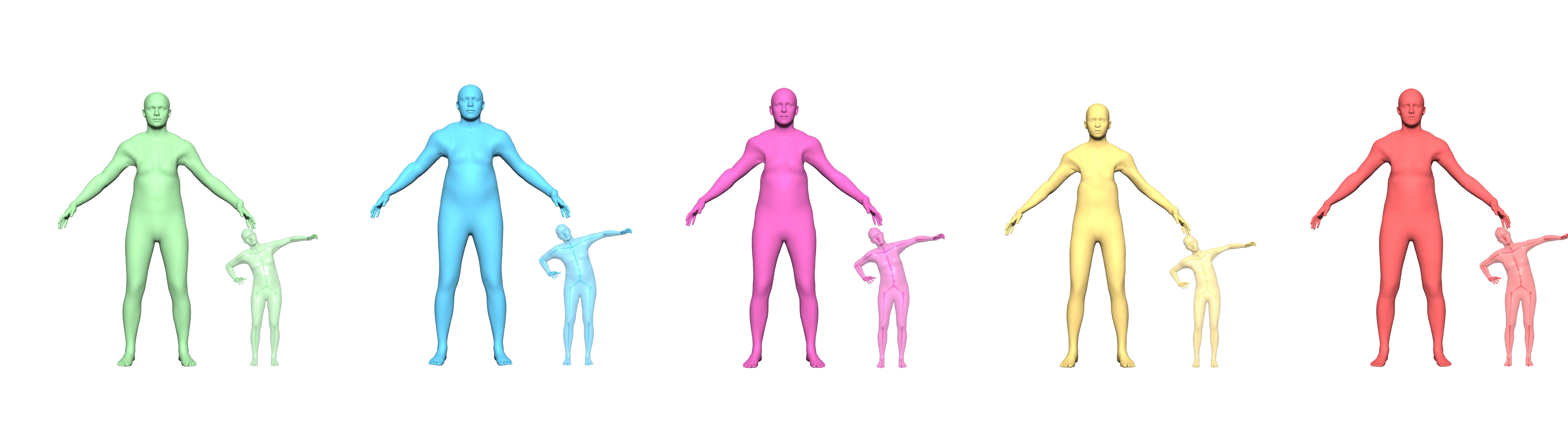}\\
\includegraphics[trim={6cm 8cm 0cm 8cm}, clip,width=\textwidth]{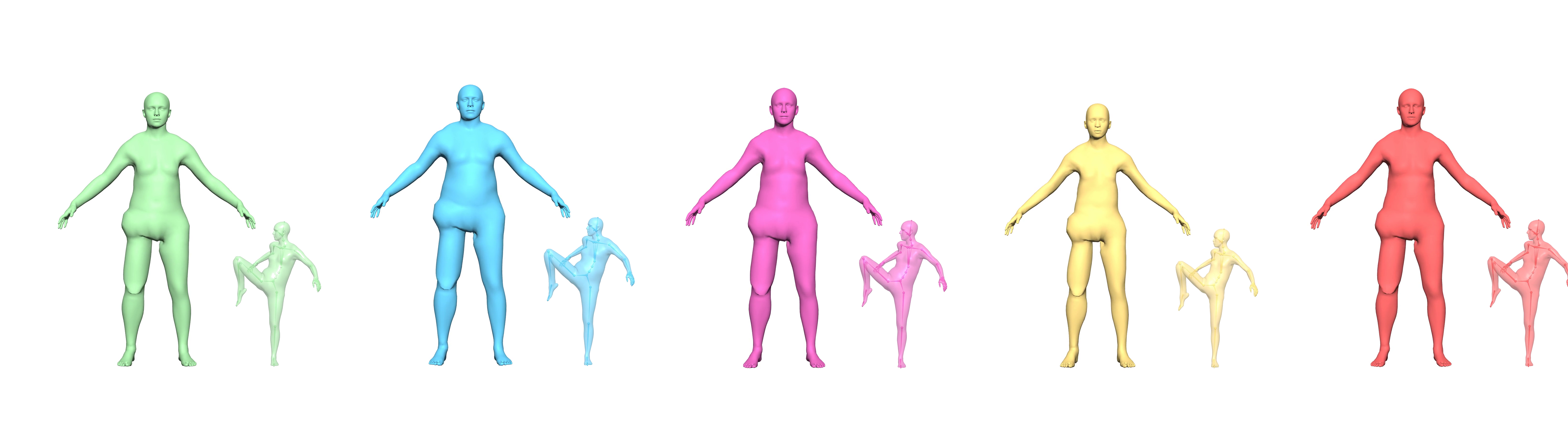}\\
\includegraphics[trim={6cm 8cm 0cm 8cm}, clip,width=\textwidth]{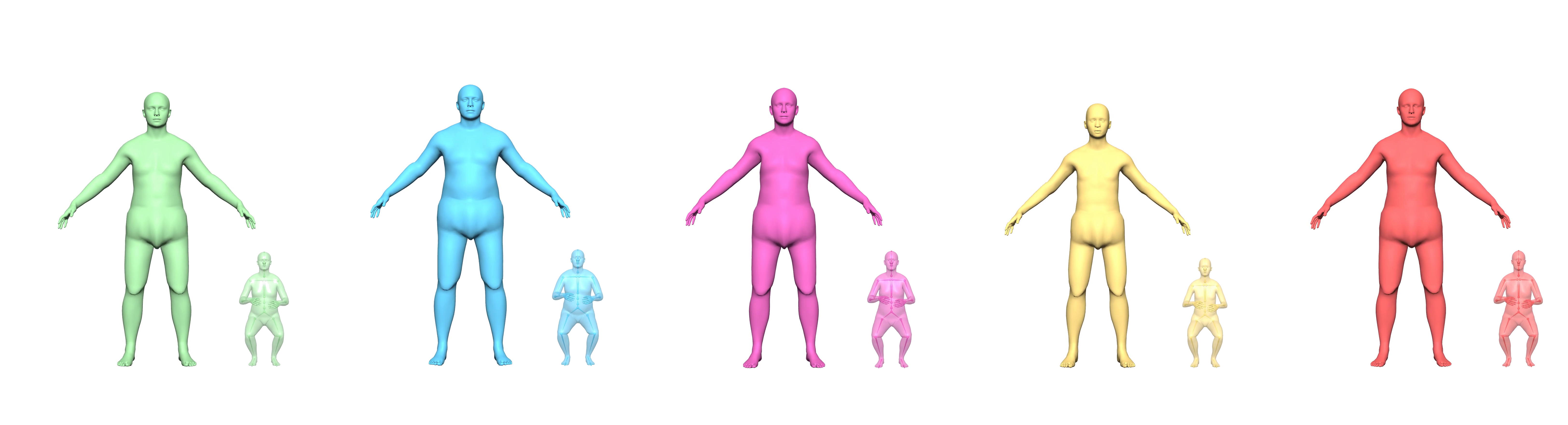}\\
\includegraphics[trim={6cm 8cm 0cm 8cm}, clip,width=\textwidth]{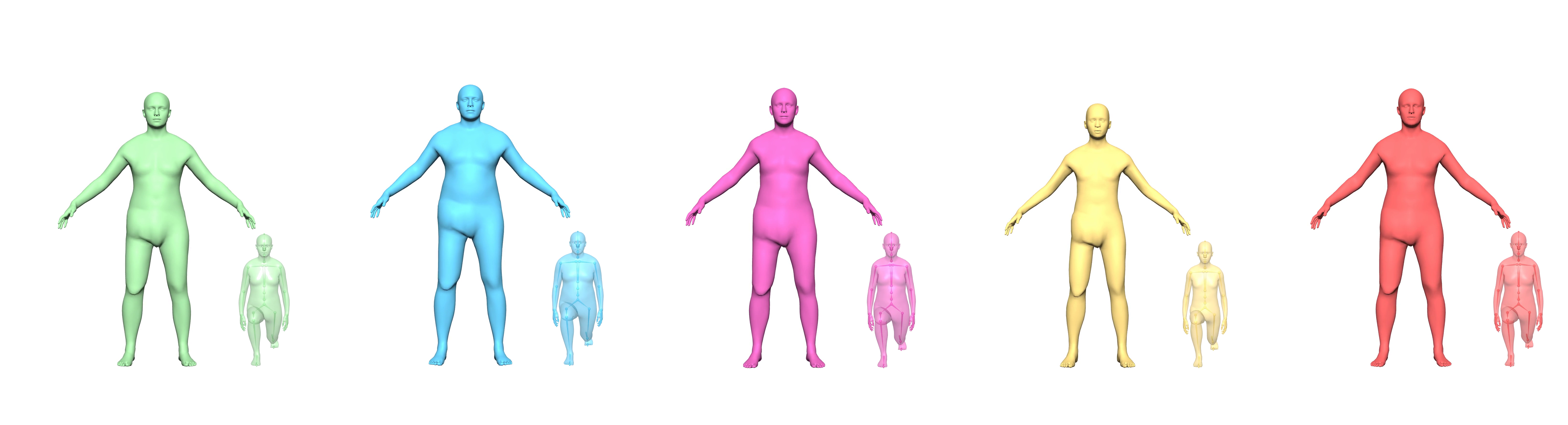}\\
\end{tabular}
\caption{%
\textbf{Unified pose correctives across identity backends.}
Each row shows a different pose; columns correspond to \name{}-Shape, MHR, SMPL, Anny, and GarmentMeasurements.
For each cell, the left mesh shows the corrective displacement applied to the canonical rest shape, and the right mesh shows the final posed result.
A single correctives model trained once on \name{}'s canonical topology produces anatomically plausible deformations for all backends.%
}
\label{fig:unified_pose_correctives}
\end{figure*}

\subsubsection{Pose-Dependent Correctives}
\label{sec:correctives}

Standard LBS produces well-known artifacts at joints undergoing large rotations
(elbow, shoulder, knee).
Some identity models ship with their own correctives (e.g.\ MHR), while others such as Anny do not.
\name{} provides a single unified correctives model that applies to \emph{all} backends, by operating on the shared canonical topology and rest pose established by the preceding abstraction layers. 

The correctives are predicted by a lightweight non-linear MLP network and applied to the rest shape before skinning:
\begin{equation}
  V_h^{\mathrm{corr}}(\beta, \theta) = V_h(\beta) + f_{\mathrm{MLP}}(\theta),
  \label{eq:corrective}
\end{equation}
where $f_{\mathrm{MLP}}$ takes as input the local joint rotations
$\{R_k(\theta)\}_{k=0}^{J-1}$ in 6D representation~\citep{zhou2019continuity}.

The MLP follows a two-stage structure inspired by MHR~\citep{MHR:2025}: joint rotations are mapped to a bank of $K = J \times C$ corrective activations ($C = 24$), which are then mapped to per-vertex displacements.
Fixed anatomical masks derived from skinning weights and geodesic distances enforce spatial locality and sparsity.
Training data is distilled from MHR by sampling ${\approx}80{,}000$ MHR posed meshes onto the \name{} topology via barycentric interpolation (\cref{sec:bary}) and pose inversion (\cref{sec:pose_inversion})---a large-scale distillation made practical by \name{}'s unified topology and pose abstraction. Fig.~\ref{fig:unified_pose_correctives} shows some examples of our unified correctives for all body models.

\subsection{Pose Abstraction}
\label{sec:pose_inversion}

The sections above describe the forward path of \name{}: given an identity and a pose, the framework produces a posed mesh in a unified representation.
A complementary operation is equally important in practice---recovering \name{} pose parameters \emph{from} an already-posed mesh.
We call this \emph{pose abstraction}: just as topology abstraction (\cref{sec:bary}) and skeletal abstraction (\cref{sec:skeleton}) unify heterogeneous body shapes into a single identity representation, pose abstraction unifies heterogeneous pose data into a single \name{} skeleton convention.
This enables motion sequences captured or generated with SMPL, MHR, Anny, or any other supported backend to be directly consumed by downstream \name{} applications without custom retargeting.
Large-scale motion datasets such as AMASS~\citep{AMASS:ICCV:2019} and SAM 3D Body~\citep{yang2026sam3dbody} thereby become natively usable through \name{}.

The core algorithmic challenge is \emph{pose inversion}: recovering per-joint rotations from posed vertex positions---the inverse of the forward kinematic and LBS pipeline described in \cref{sec:lbs}.
We describe the pose inversion algorithm below.

\paragraph{Multi-topology input}
Posed vertices from any supported mesh topology---not only the native \name{} mesh---are accepted as input.
When the input topology differs from \name{}'s canonical mesh, the same barycentric transfer used in the forward path (\cref{sec:bary}) first maps the input vertices to the \name{} topology.
From this point onward, the inversion algorithm operates entirely in \name{}'s canonical vertex and skeleton space.

\paragraph{Initialization via skeleton transfer}
Pose inversion begins with the same Kabsch-based skeleton fitting procedure described in \cref{sec:skeleton}: a single-pass RBF joint regression followed by Procrustes alignment provides an initial world-space rotation estimate for each joint.
This initialization is already a reasonable approximation of the target pose and can serve as a standalone fast solver when only coarse pose recovery is needed.

\paragraph{Iterative inverse-LBS refinement}
Starting from the skeleton-transfer initialization, the algorithm refines joint rotations level-by-level in parent-to-child order.
For joint $k$, the skinned vertex positions are decomposed into a subtree contribution (vertices predominantly influenced by $k$ and its descendants) and a non-subtree contribution from already-solved ancestor joints.
The ancestor contribution is subtracted from the observed posed vertices, isolating the local deformation attributable to joint $k$ alone.
A Procrustes alignment (\cref{eq:procrustes}) is then solved for the local rotation.

\paragraph{Newton-Schulz orthogonalization}
The standard Kabsch algorithm computes the nearest rotation matrix from the cross-covariance matrix $H = A^T B$ via SVD: $R = U V^T$ where $H = U \Sigma V^T$.
However, when the point cloud contributing to a joint's covariance is near-coplanar---as commonly occurs at body parts such as clavicles---the smallest singular value $\sigma_3$ approaches zero and the corresponding singular vector becomes ill-defined.
Under these conditions, small perturbations in the input vertices can flip the sign of a singular vector between consecutive frames, causing a discontinuous $180°$ rotation jump (``shoulder popping'').
To avoid this instability, our iterative refinement replaces SVD with Newton-Schulz orthogonalization~\citep{kovarik1970newton}, which computes the polar factor of $H$ via the fixed-point iteration
\begin{equation}
    R_{i+1} = \tfrac{1}{2} R_i (3I - R_i^T R_i), \quad R_0 = H / \|H\|_\infty,
\label{eq:newton_schulz}
\end{equation}
where $\|H\|_\infty$ is the infinity norm (maximum absolute row sum) that guarantees convergence.
Because this iteration refines the rotation estimate \emph{continuously} from its current value rather than decomposing and reassembling singular vectors, it is immune to the sign-flipping discontinuity inherent in SVD for near-degenerate covariance matrices.

\paragraph{Hierarchical scheduling}
The refinement schedule mirrors the skeletal hierarchy: body joints are solved first, followed by optional finger joint refinement, and a final global pass covers all joints simultaneously.
This coarse-to-fine schedule ensures that large-scale body motion is resolved before fine-grained finger articulation, which would otherwise be contaminated by uncorrected upstream errors.

\paragraph{Optional autograd refinement}
For applications that require higher accuracy at the cost of throughput, an optional gradient-based refinement stage is provided.
Joint rotations are parameterized as continuous 6D vectors~\citep{zhou2019continuity} and optimized with Adam by backpropagating through the full FK+LBS computation (\cref{eq:lbs}).
This stage must be warm-started from the analytical result: the FK+LBS objective is highly non-convex, and naïve optimization from the bind pose fails to converge, settling into a local minimum with entirely incorrect limb placement (\cref{sec:eval_pose}).
With the analytical initialization, autograd refinement converges rapidly and can further reduce error at extremities (hands, feet, head) by optimizing through the full kinematic chain with per-vertex loss weighting.
The analytical solver alone achieves approximately 1{,}200 frames per second on an NVIDIA RTX 5000 Ada GPU, while the autograd path runs at 16--18 FPS for 100 optimizer steps, making each mode suitable for different points on the speed--accuracy tradeoff curve.

%% file: sections/04_experiments.tex
\section{Evaluation}
\label{sec:eval}

We evaluate \name{} along four dimensions: topology abstraction fidelity (\cref{sec:eval_geo}), pose inversion accuracy (\cref{sec:eval_pose}), runtime performance (\cref{sec:eval_perf}), and cross-model shape-space comparison (\cref{sec:eval_shape}).

\subsection{Topology Abstraction Fidelity}
\label{sec:eval_geo}

The topology transfer is the first stage of the pipeline and any error here propagates to all downstream tasks.
Tab.~\ref{tab:geo_error} reports per-vertex transfer errors for each backend over 100 diverse identities.
For each \name{} vertex, we query the closest point on the native source mesh surface and report the $L_2$ distance.
This measures the geometric information loss introduced by the barycentric interpolation, without conflating it with unit convention or alignment artifacts.

\begin{table}[t]
\centering
\caption{%
\textbf{Topology abstraction fidelity across backends.}
Closest-point-to-mesh distance (mm) between the \name{} rest shape and the native source mesh, measured over 100 diverse identities per backend.
``Wrap'' is the baseline registration error of the pre-registered \name{} wrap mesh against the source neutral mesh.
Vertices in facial inner geometry (eye bags, mouth bag) and between-toes regions---which have no correspondence in most source topologies---are excluded.%
}
\label{tab:geo_error}
\begin{tabular}{lrrrrrr}
\toprule
\textbf{Backend} & \textbf{Src.~Verts} & \textbf{Mean (mm)} & \textbf{Std (mm)} & \textbf{P95 (mm)} & \textbf{Wrap Mean} & \textbf{Wrap P95} \\
\midrule
\name{} native    & --     & 0.0 & 0.0 & 0.0 & -- & -- \\
SMPL           & 6{,}890  & 0.12 & 0.45 & 0.71 & 0.12 & 0.74 \\
SMPL-X         & 10{,}475 & 0.06 & 0.22 & 0.45 & 0.06 & 0.47 \\
Anny           & 13{,}718 & 0.01 & 0.12 & 0.01 & 0.01 & 0.01 \\
MHR            & ${\sim}$18k & 0.40 & 0.73 & 1.49 & 0.31 & 1.28 \\
\bottomrule
\end{tabular}
\end{table}

The \name-native backend incurs zero error by construction.
SMPL and SMPL-X achieve sub-millimeter mean errors (0.12\,mm and 0.06\,mm respectively), and their transfer errors closely match the wrap baseline, confirming that the barycentric interpolation introduces negligible additional distortion beyond the one-time mesh registration.
Anny achieves near-zero error (0.01\,mm mean), reflecting a particularly clean wrap registration.
MHR shows a slightly higher mean error (0.40\,mm), reflecting its denser mesh and more complex geometry, but remains well below 1\,mm; the modest increase over the wrap baseline (0.31\,mm) indicates that the topology transfer generalizes well across the MHR shape space.
All P95 errors stay below 1.5\,mm.

\begin{figure*}[t]
\centering
\includegraphics[width=\textwidth]{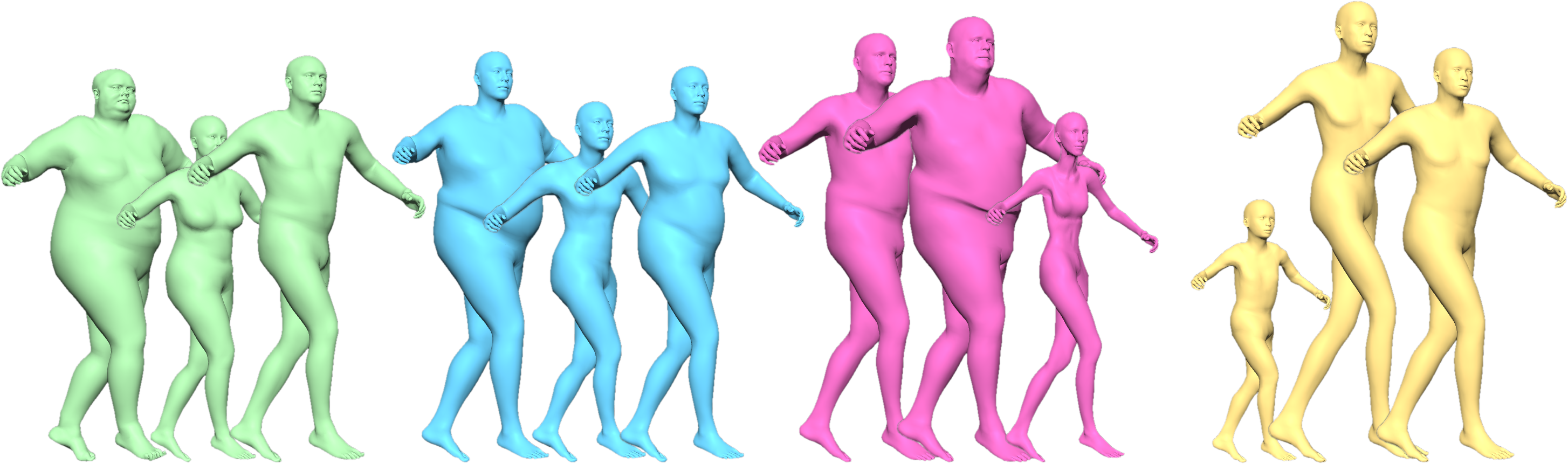}
\caption{%
\textbf{Shape diversity across backends driven by a single pose.}
Three sampled identities per backend---\name-Shape (green), MHR (blue), SMPL (pink), and Anny (yellow)---all driven by the same skeletal pose.
Despite originating from entirely different generative models, all bodies share the same pose interpretation, illustrating the plug-and-play nature of \name identity-pose decoupling.%
}
\label{fig:diversity}
\end{figure*}

Fig.~\ref{fig:diversity} shows the shape diversity achieved across all backends using identical pose parameters, demonstrating that the unified pipeline faithfully represents the shape characteristics of each source model.

\subsection{Pose Inversion Accuracy}
\label{sec:eval_pose}

Posed meshes are the common interface across heterogeneous body models, making mesh-to-joint-angle inversion the key operation for pose abstraction.
Tab.~\ref{tab:pose_inv} reports pose inversion accuracy and throughput for both solvers described in \cref{sec:pose_inversion}.
We evaluate on the full AMASS dataset~\citep{AMASS:ICCV:2019} (344 subjects, 2{,}265 motions, 40.3 hours, 19.8M frames).
Errors are per-vertex $L_2$ distances between the original SMPL-X posed mesh and the \name{} reconstruction driven by the recovered rotations.

\begin{table}[t]
\centering
\caption{%
\textbf{Pose inversion accuracy and throughput on AMASS.}
Per-vertex reconstruction error (mm) and throughput (frames/sec) on an NVIDIA A100 GPU.
``Skel.\ transfer'' is the raw skeleton-fitting initialization with no iterative refinement.
``Analytical'' adds inverse-LBS refinement with Newton-Schulz orthogonalization (body=2, full=1).
``Autograd FK'' optimizes 6D rotation parameters through FK+LBS with Adam (100 iterations); ``no init'' starts from the bind pose, ``w/ init'' warm-starts from the skeleton transfer result.
``Analytical + Autograd'' warm-starts autograd from the analytical result (10 iterations).%
}
\label{tab:pose_inv}
\begin{tabular}{lrrrr}
\toprule
\textbf{Method} & \textbf{Mean (mm)} & \textbf{Median (mm)} & \textbf{Max (mm)} & \textbf{FPS} \\
\midrule
Skel.\ transfer only         & 16.5 & 13.9 & 80.1 & 17{,}393 \\
Analytical                    &  5.3 &  3.2 & 88.1 &    882 \\
Autograd FK (no init)         & 501.8 &  479.4 & 1354.2 &     79 \\
Autograd FK (w/ init)         &  4.1 &  2.1 & 81.5 &     78 \\
Analytical + Autograd (10)    & 7.8 &  6.4 & 88.6 &    435 \\
\bottomrule
\end{tabular}
\end{table}

\begin{figure}[t]
\centering
\begin{tabular}{@{}cc@{}}
\includegraphics[width=0.42\linewidth]{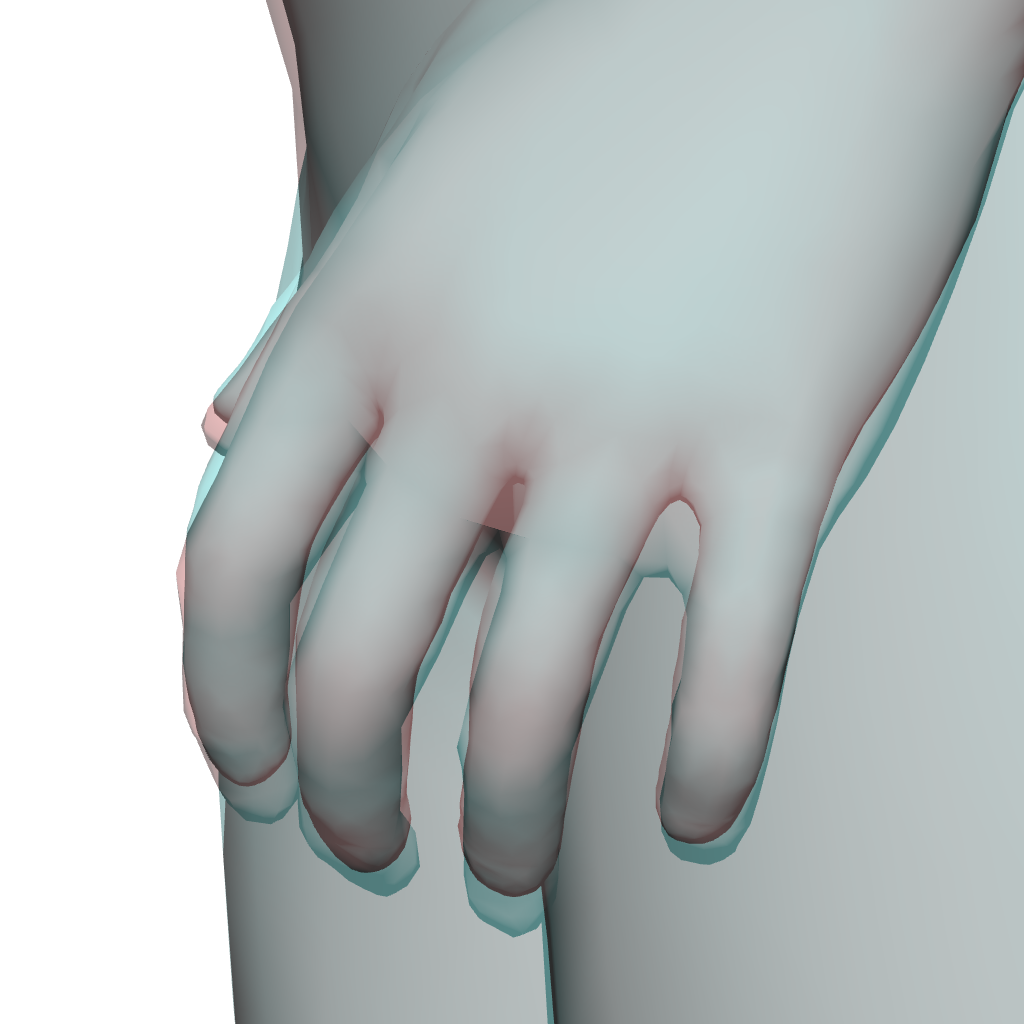} &
\includegraphics[width=0.42\linewidth]{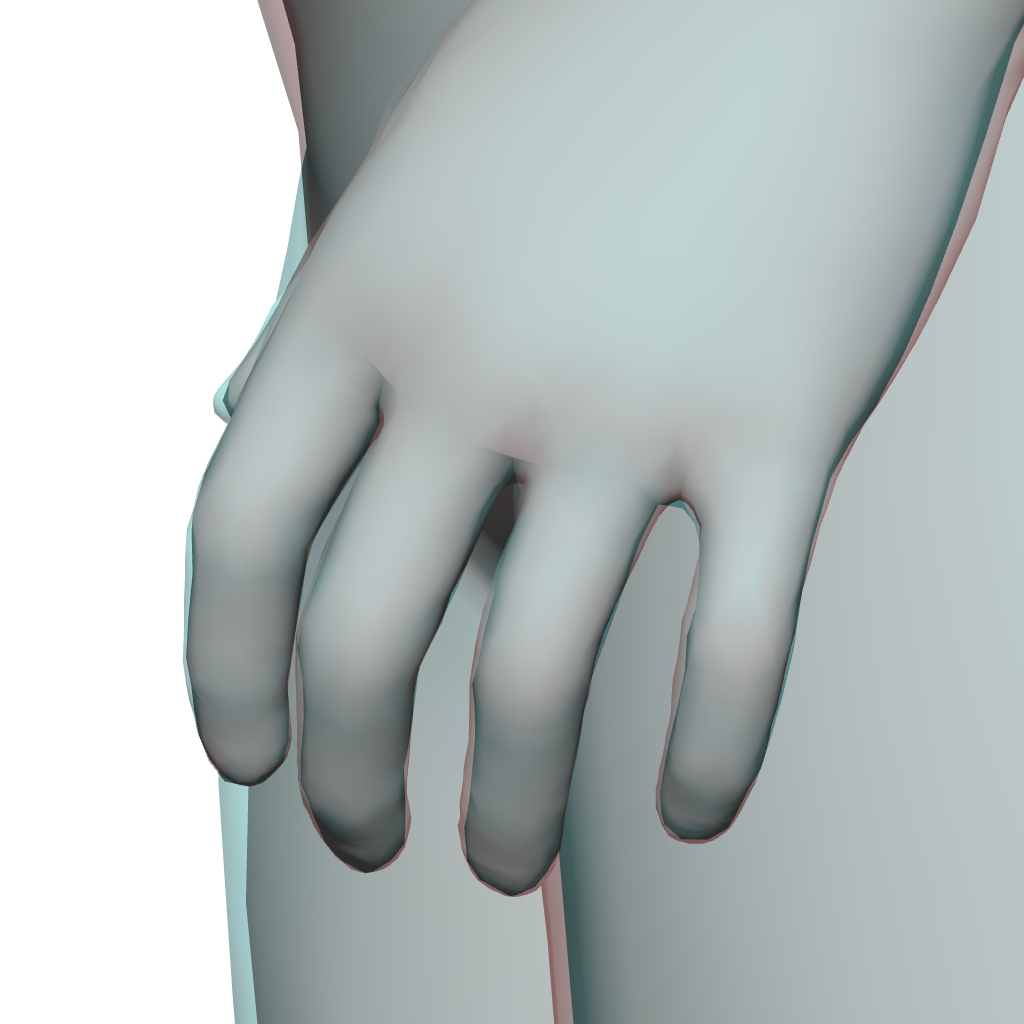} \\
\small (a) Analytical only &
\small (b) Analytical + Autograd FK \\
\end{tabular}
\caption{%
\textbf{Hand zoom: analytical vs.\ autograd FK refinement.}
MHR backend on SAM 3D Body (teal = ground truth, red = \name{} reconstruction).
(a)~The analytical solver is near-optimal globally but leaves residual misalignment at the fingertips.
(b)~Adding 100 autograd FK iterations redistributes error toward the body trunk (visible as a slightly increased offset in the background), achieving a tight overlay at the fingers.%
}
\label{fig:hand_zoom}
\end{figure}

\begin{figure}[t]
\centering
\begin{tabular}{@{}ccc@{}}
\includegraphics[width=0.32\linewidth]{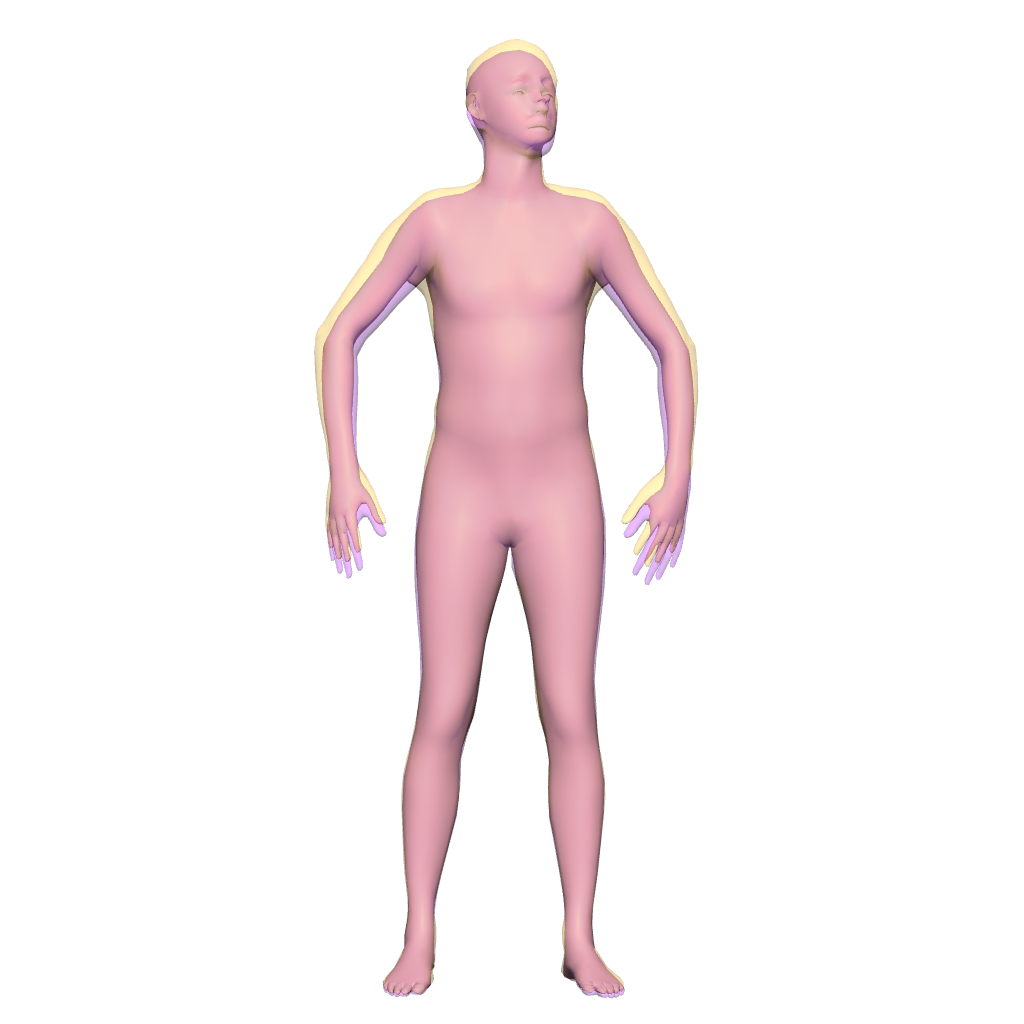} &
\includegraphics[width=0.32\linewidth]{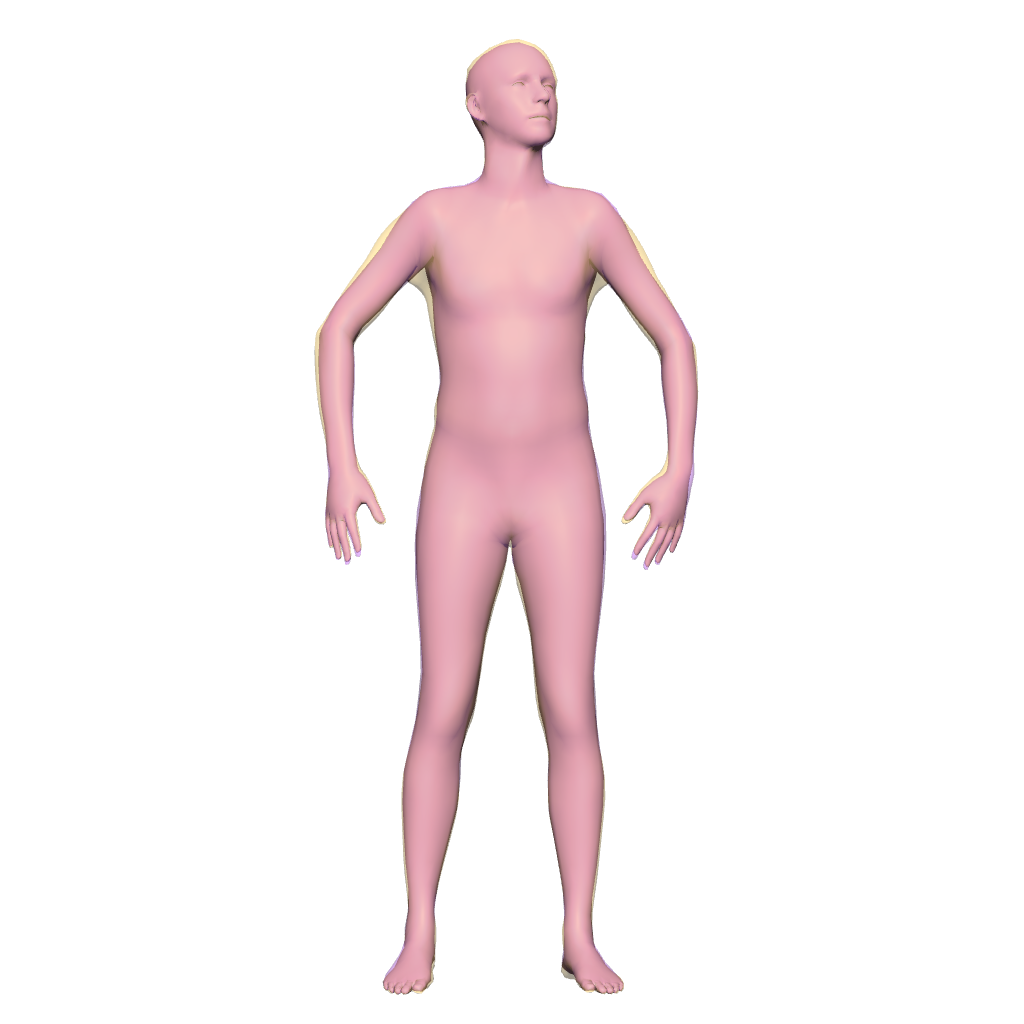} &
\includegraphics[width=0.32\linewidth]{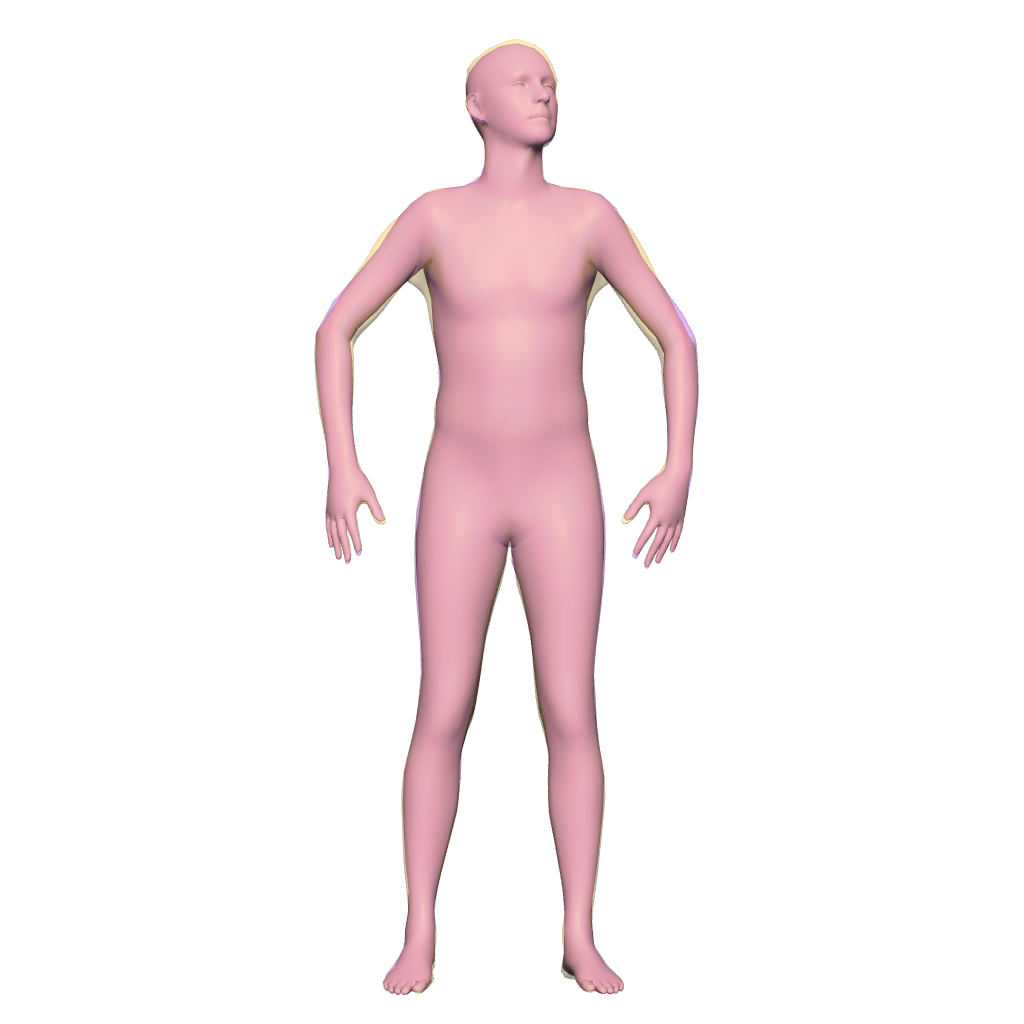} \\
\small (a) Skel.\ transfer only &
\small (b) Analytical &
\small (c) Analytical + Autograd FK \\[4pt]
\includegraphics[width=0.32\linewidth]{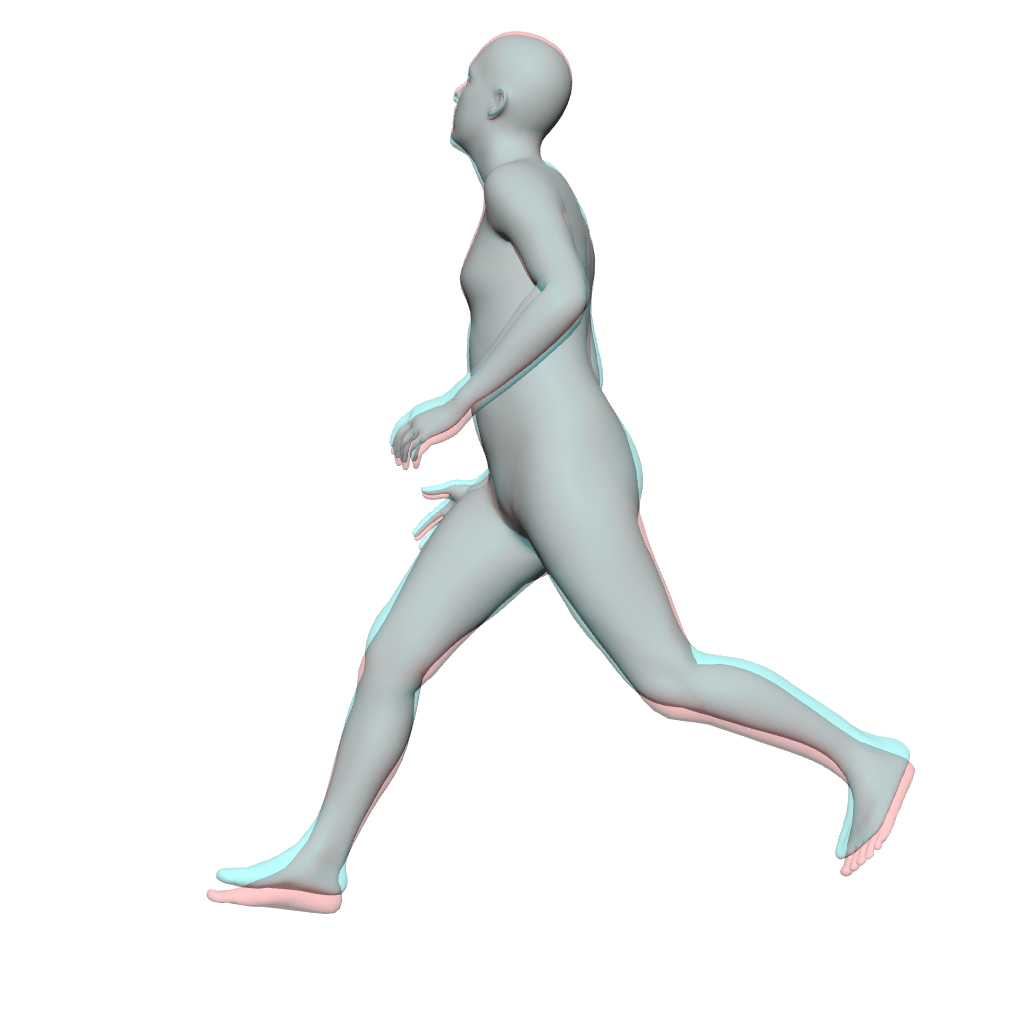} &
\includegraphics[width=0.32\linewidth]{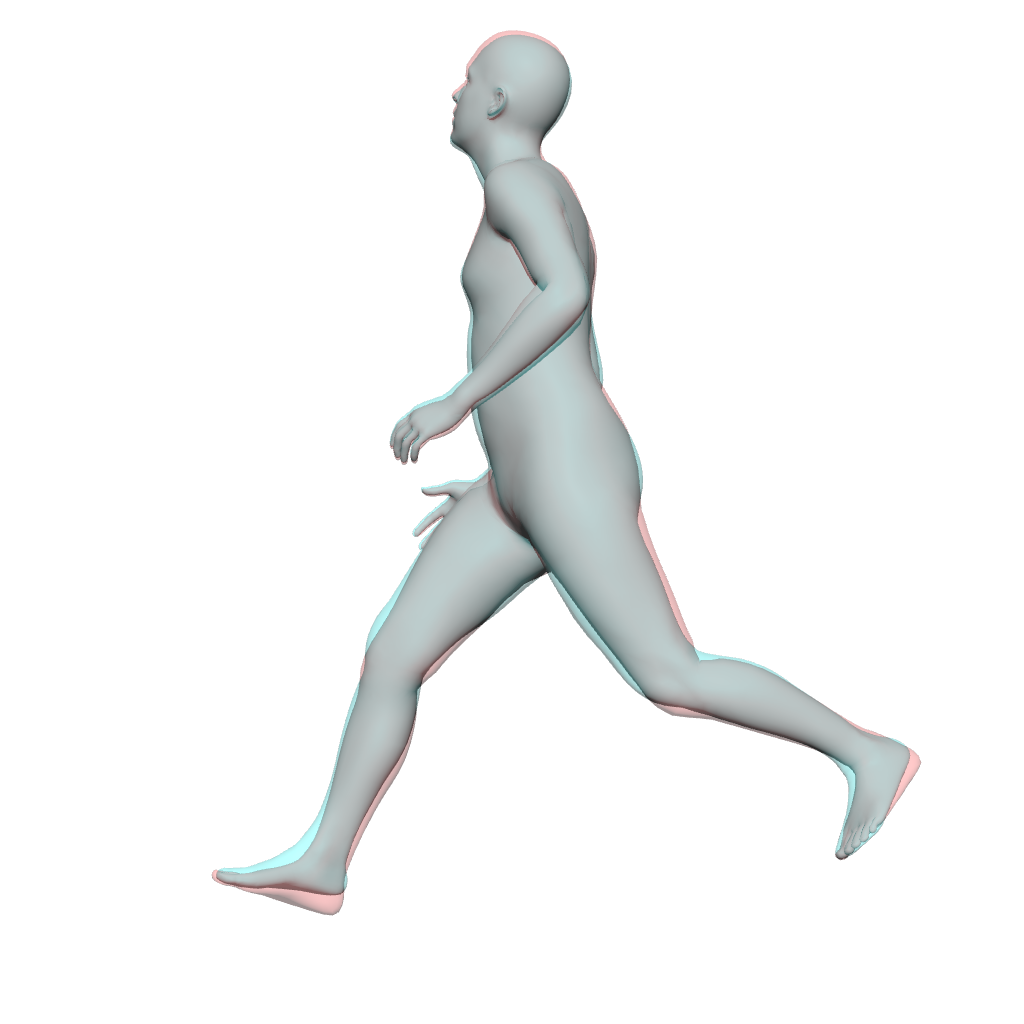} &
\includegraphics[width=0.32\linewidth]{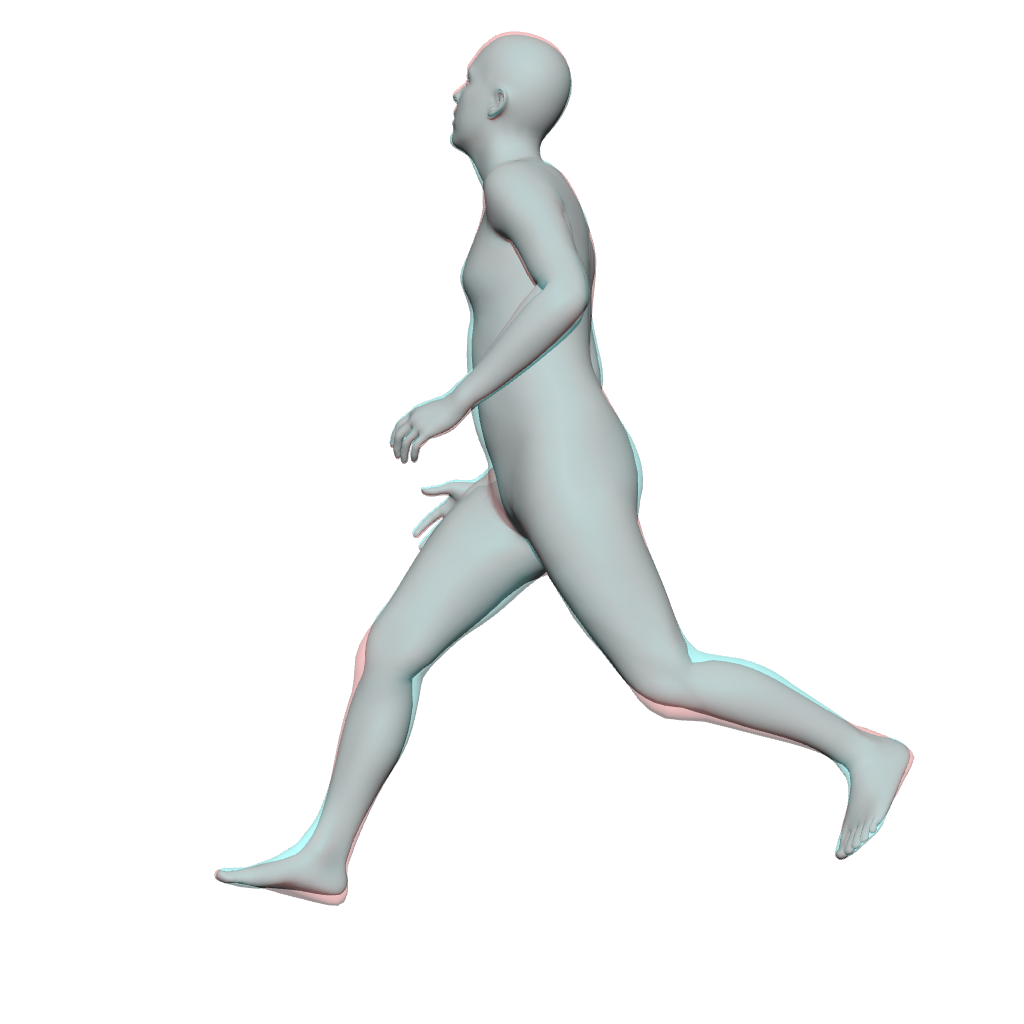} \\
\small (d) Skel.\ transfer only &
\small (e) Analytical &
\small (f) Analytical + Autograd FK \\
\end{tabular}
\caption{%
\textbf{Pose inversion quality across methods.}
Top row: SMPL backend (yellow = ground truth, purple = \name{} reconstruction).
Bottom row: MHR backend on SAM 3D Body (teal = ground truth, red = \name{} reconstruction).
(a,\,d)~Skeleton transfer alone provides a coarse estimate with visible misalignment at the extremities.
(b,\,e)~Analytical refinement with Newton-Schulz orthogonalization closely tracks the ground truth.
(c,\,f)~Adding autograd FK refinement achieves the tightest overlay, reducing residual error at the feet and hands.%
}
\label{fig:pose_inv}
\end{figure}

Fig.~\ref{fig:pose_inv} qualitatively compares the three stages on SMPL and MHR backends.
The skeleton transfer initialization alone (\cref{sec:skeleton}) provides a coarse but fast pose estimate (16.5\,mm mean at 17{,}393 FPS), suitable for applications where speed dominates accuracy requirements.
The analytical solver refines this to 5.3\,mm mean error at 882 FPS via iterative inverse-LBS with Newton-Schulz orthogonalization.
The autograd FK solver reaches 4.1\,mm mean error by optimizing through the full FK+LBS chain, but \emph{only when warm-started} from the skeleton transfer initialization; without initialization (starting from the bind pose), 100 Adam iterations fail to converge (501.8\,mm mean error), demonstrating that the initialization is critical.
The two solvers are complementary.
The analytical path is fast and produces a near-optimal result in terms of global $L_2$ error across all vertices.
The autograd FK path, by contrast, optimizes through the full FK+LBS chain and supports per-vertex loss weighting on extremities (hands, feet, head), giving explicit control over where the solver concentrates its effort.
Tab.~\ref{tab:extremity} breaks down per-vertex errors by body region on 200 SAM 3D Body~\citep{yang2026sam3dbody} frames (MHR backend).
Autograd FK refinement (100 iterations, warm-started from the analytical result) reduces hand error by 57\% (4.7\,$\to$\,2.0\,mm), foot error by 29\% (8.2\,$\to$\,5.8\,mm), and head error by 30\% (6.9\,$\to$\,4.8\,mm), while body trunk error decreases slightly (16.8\,$\to$\,15.3\,mm)---the optimizer redistributes error away from the extremities and onto the trunk, which has more vertices to absorb it.
Fig.~\ref{fig:hand_zoom} visualizes this trade-off on a hand close-up: the autograd result achieves a tight overlay at the fingertips, at the cost of a slightly increased offset on the body visible in the background.

\begin{table}[t]
\centering
\caption{%
\textbf{Per-region pose inversion error (MHR, 200 SAM 3D Body frames).}
Mean per-vertex $L_2$ error (mm) by body region.
``Analytical'' = body=2, full=1.
``+ Autograd'' adds 100 Adam iterations warm-started from the analytical result.%
}
\label{tab:extremity}
\begin{tabular}{lrrrrr}
\toprule
\textbf{Method} & \textbf{All} & \textbf{Body} & \textbf{Hands} & \textbf{Feet} & \textbf{Head} \\
\midrule
Analytical           & 8.8 & 16.8 & 4.7 & 8.2 & 6.9 \\
+ Autograd (100)     & 6.6 & 15.3 & 2.0 & 5.8 & 4.8 \\
\midrule
\textbf{Reduction}   & 25\% & 9\% & \textbf{57\%} & 29\% & 30\% \\
\bottomrule
\end{tabular}
\end{table}

\begin{figure}[t]
\centering
\begin{tabular}{@{}cccc@{}}
\includegraphics[width=0.24\linewidth]{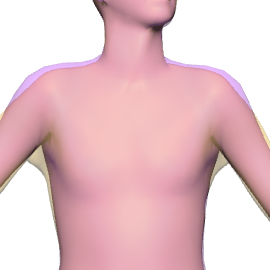} &
\includegraphics[width=0.24\linewidth]{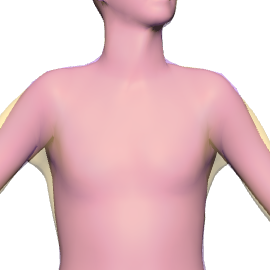} &
\includegraphics[width=0.24\linewidth]{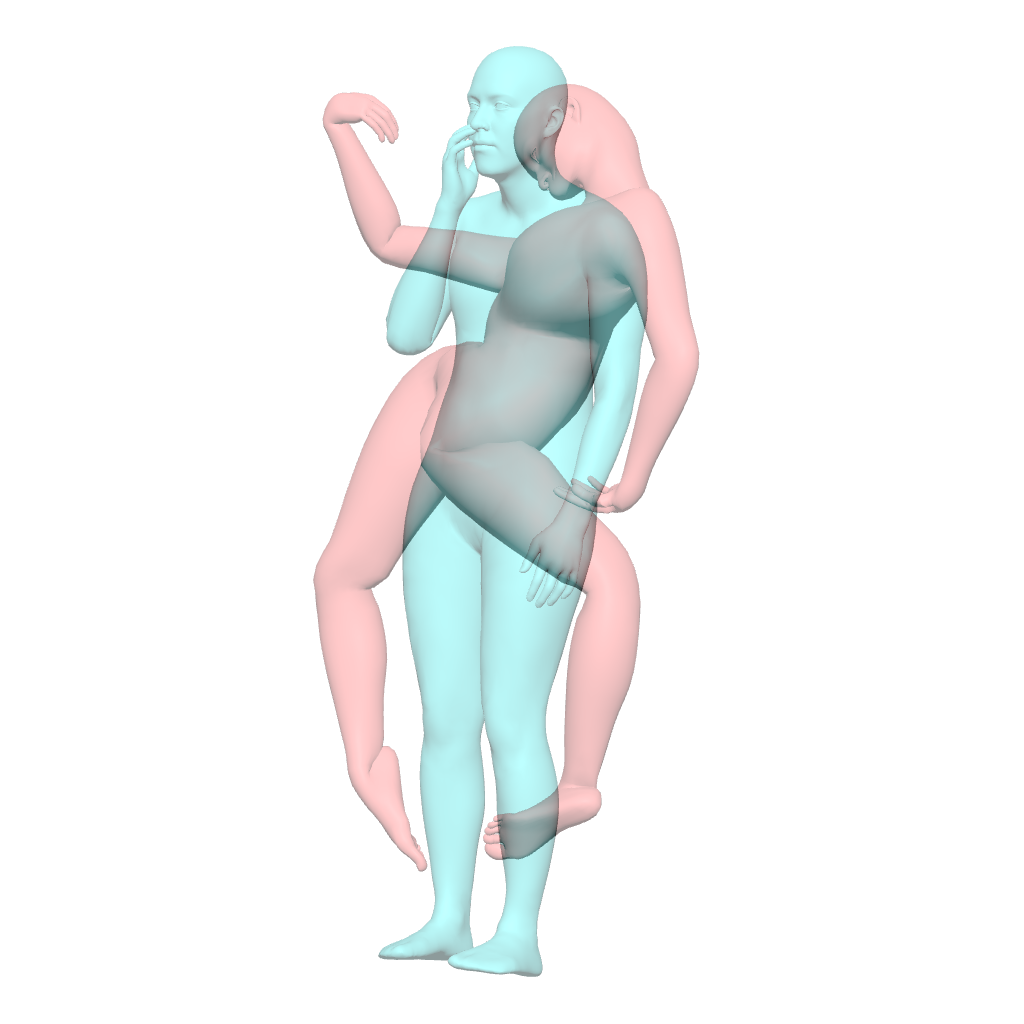} &
\includegraphics[width=0.24\linewidth]{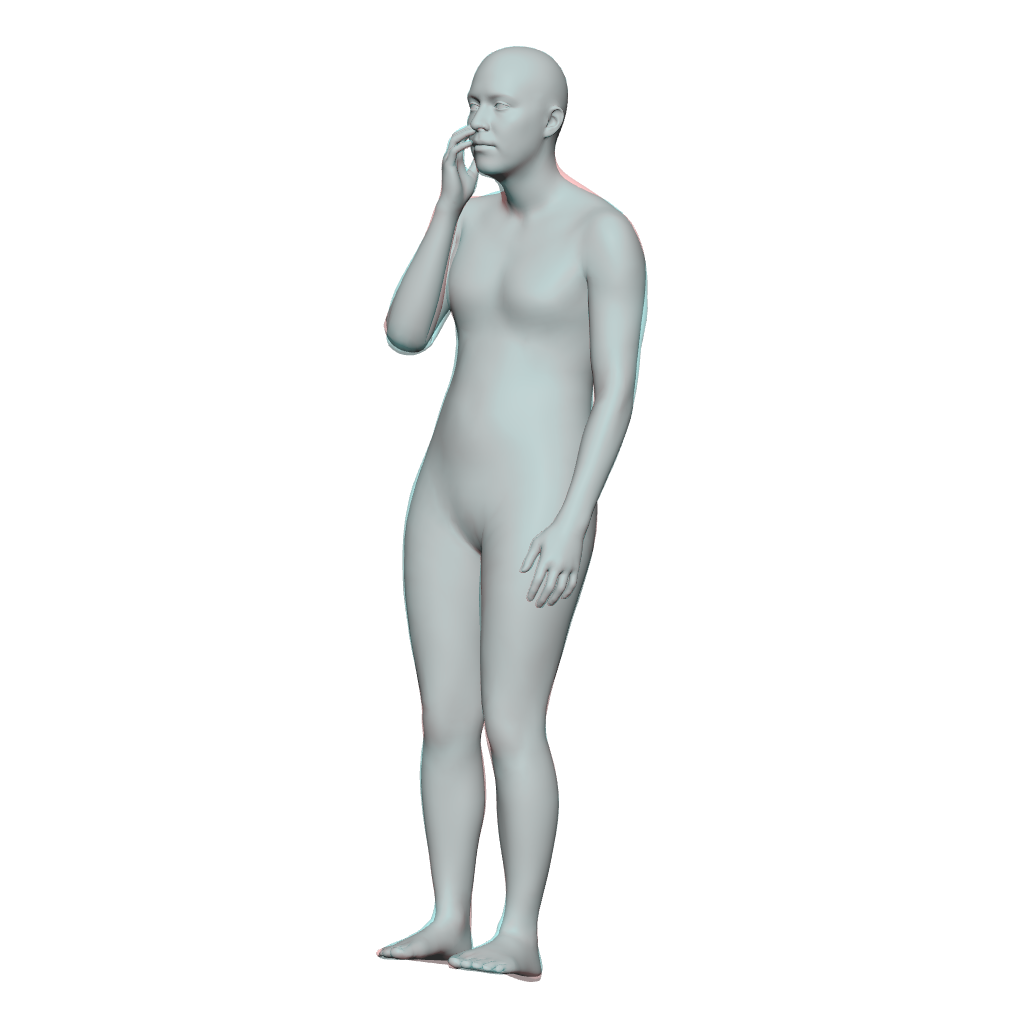} \\
\small (a) SVD & \small (b) Newton-Schulz &
\small (c) Autograd (no init) & \small (d) Analytical \\
\end{tabular}
\caption{%
\textbf{Pose inversion ablations.}
(a,\,b)~Shoulder zoom from a single SMPL frame (yellow = GT, purple = \name{}).
SVD-based Kabsch alignment exhibits shoulder popping due to a singular-vector sign flip; Newton-Schulz produces a smooth result.
(c,\,d)~MHR on SAM 3D Body (teal = GT, red = \name{}).
Without initialization, 100 autograd iterations converge to an incorrect local minimum; the analytical solver recovers the pose correctly.%
}
\label{fig:pose_inv_ablations}
\end{figure}

\paragraph{Effect of Newton-Schulz orthogonalization}
Fig.~\ref{fig:pose_inv_ablations}(a,\,b) compares the analytical solver using standard SVD-based Kabsch alignment against the Newton-Schulz variant described in \cref{sec:pose_inversion}.
The crops are taken from the shoulder region of the same SMPL frame, where the contributing vertex cloud is near-coplanar.
With SVD, the near-zero third singular value causes a sign flip in the rotation solution, producing a visible discontinuous offset at both shoulders (``shoulder popping'').
Newton-Schulz orthogonalization avoids this instability by iteratively refining the rotation estimate without decomposing singular vectors, yielding a smooth and accurate result.
To quantify the temporal effect, we measure frame-to-frame shoulder-region vertex error change across the full 1{,}606-frame sequence: SVD exhibits a peak error oscillation of 1.6\,mm/frame at the shoulders, compared to 0.8\,mm/frame for Newton-Schulz---a $2{\times}$ improvement in temporal stability.

\paragraph{Importance of initialization}
Fig.~\ref{fig:pose_inv_ablations}(c,\,d) illustrates why initialization is critical for gradient-based pose inversion: on a simple standing pose from SAM 3D Body, 100 Adam iterations without initialization converge to a local minimum with entirely incorrect limb placement, while the analytical solver recovers the pose in a single pass with near-perfect overlay.

\subsection{Runtime Performance}
\label{sec:eval_perf}

\name{} integrates directly into large-scale foundation model training loops, so the forward pass must be highly optimized.
Tab.~\ref{tab:perf} reports throughput and latency across batch sizes and execution modes.
Measurements were conducted on a single NVIDIA A100 80\,GB GPU (Warp path) and a 32-core AMD EPYC 7763 CPU (PyTorch path), with the mid-resolution \name{} mesh and a \name-native identity backend (which skips the topology abstraction step).

The skeleton fitting step (RBF regression + Kabsch) takes under 1.5\,ms on GPU regardless of batch size, demonstrating that the pre-factored sparse matrix approach scales efficiently.
The Warp GPU path achieves over 7{,}000 meshes/sec at batch size 128.
Adding a topology abstraction step (SMPL or MHR backend) incurs approximately 0.3--0.8\,ms additional latency on GPU, which is negligible relative to total pipeline cost.

\begin{table}[t]
\centering
\caption{%
\textbf{Runtime performance.}
Throughput (meshes/sec) and per-call latency (ms) of the \name{} forward pass.
Breakdown shows skeleton fitting cost (RBF regression + Kabsch) vs.\ total forward pass.
Warp = GPU path (NVIDIA Warp LBS kernel); PyTorch = CPU dense path.
Identity backend: \name-native (no topology abstraction overhead).%
}
\label{tab:perf}
\begin{tabular}{lrrrr}
\toprule
\textbf{Mode} & \textbf{Batch} & \textbf{Skel.\,(ms)} & \textbf{Total\,(ms)} & \textbf{Meshes/sec} \\
\midrule
Warp (GPU)     &  1  & 0.8 &  2.1 &   476 \\
Warp (GPU)     &  8  & 0.9 &  3.4 & 2{,}353 \\
Warp (GPU)     & 32  & 1.1 &  6.8 & 4{,}706 \\
Warp (GPU)     & 128 & 1.4 & 18.2 & 7{,}033 \\
PyTorch (CPU)  &  1  & 3.2 & 12.1 &    83 \\
PyTorch (CPU)  &  8  & 4.1 & 38.7 &   207 \\
PyTorch (CPU)  & 32  & 5.9 &148.0 &   216 \\
\bottomrule
\end{tabular}
\end{table}

\subsection{Cross-Model Shape-Space Comparison}
\label{sec:eval_shape}

One of the key benefits of \name's abstraction layers is principled cross-model comparison. We demonstrate this by evaluating four PCA shape models on 33 held-out body scans from an independent capture pipeline without overlap with any model's PCA training data. This is typically infeasible without a unifying framework like \name, since each model defines its own topology, rest pose, and deformation parameters.
For each model, every scan is transferred to the model's native mesh topology via barycentric interpolation (\cref{sec:bary}), reposed to the model's canonical rest pose via pose inversion (\cref{sec:pose_inversion}), and projected onto the model's PCA basis at full capacity.
Tab.~\ref{tab:pca_crossmodel} reports per-vertex reconstruction error.

\begin{table}[t]
\centering
\caption{%
\textbf{Cross-model PCA reconstruction on 33 held-out body scans.}
Per-vertex $L_2$ distance (mm) between the unposed scan and the PCA reconstruction at each model's full component count $K$.
All models are evaluated through the same \name{} pipeline, differing only in the target topology, rest pose, and PCA basis.%
}
\label{tab:pca_crossmodel}
\begin{tabular}{lrrrr}
\toprule
\textbf{Model} & \textbf{$K$} & \textbf{Mean (mm)} & \textbf{Median (mm)} & \textbf{P95 (mm)} \\
\midrule
SMPL                & 10  & 14.11 & 12.31 & 30.49 \\
GarmentMeasurements & 15  & 11.81 & 10.67 & 24.18 \\
\name-Shape (Ours)  & 128 &  5.82 &  4.81 & 13.60 \\
SMPL-X              & 300 &  5.45 &  4.34 & 12.97 \\
\bottomrule
\end{tabular}
\end{table}

\begin{figure*}[t]
\centering
\includegraphics[width=\textwidth]{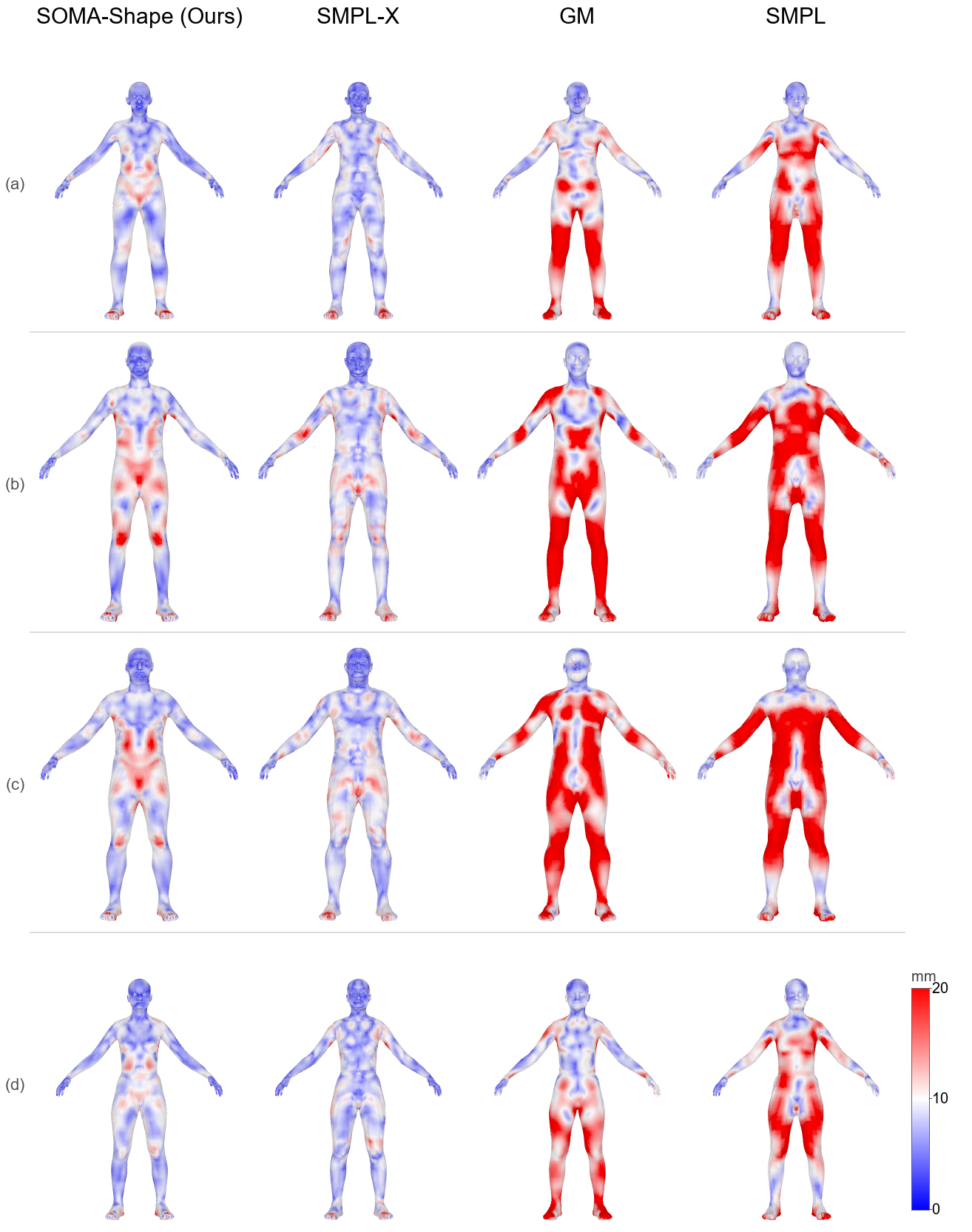}
\caption{%
\textbf{Cross-model PCA reconstruction error on held-out body scans.}
\name{}'s topology and pose abstraction layers enable a principled cross-model comparison.
Each column shows a different shape model's mesh topology, all rendered in \name{} A-pose; each row (a--d) is the \emph{same} identity.
Color encodes per-vertex $L_2$ error (blue\,=\,0\,mm, white\,=\,10\,mm, red\,$\geq$\,20\,mm).
SMPL (10 components) and GM (GarmentMeasurements, 15) show widespread residual, while \name-Shape (128) and SMPL-X (300) achieve low error across the body surface.%
}
\label{fig:pca_crossmodel}
\end{figure*}

SMPL's 10-component basis captures coarse body proportions but leaves a 14\,mm mean residual, consistent with its limited shape dimensionality.
GarmentMeasurements (15 components) reduces this to 12\,mm.
\name-Shape achieves 5.8\,mm mean with 128 components, closely matching SMPL-X's 5.5\,mm at 300 components---demonstrating competitive expressiveness with fewer than half the parameters.
Fig.~\ref{fig:pca_crossmodel} visualizes the per-vertex reconstruction error across models for representative scans: SMPL and GarmentMeasurements show widespread residual (red), while \name-Shape and SMPL-X achieve low error over most of the body surface (blue).

%% file: sections/05_conclusion.tex
\section{Conclusion}
\label{sec:conclusion}

We have presented \name{}, a unified framework that decouples identity representation from pose parameterization across heterogeneous parametric body models.
By mapping all supported backends to a single canonical mesh topology and rig, \name{} reduces the $O(M^2)$ per-pair adapter problem to $O(M)$ single-backend connectors, enabling practitioners to freely mix identity sources and pose data at inference time.
Three abstraction layers---mesh topology, skeleton, and pose---unify heterogeneous body shapes into a single identity representation, adapt the skeleton to arbitrary identities and pose them with unified corrective deformations shared across all backends, and recover unified skeleton rotations directly from posed vertices without custom retargeting.
The entire pipeline is fully differentiable, GPU-accelerated, and requires no per-model training or iterative optimization.
Evaluation across multiple backends and 100 diverse identities demonstrates sub-millimeter mean topology transfer errors on body vertices, sub-centimeter pose inversion accuracy at over 300 FPS on a laptop GPU, and forward-pass throughput exceeding 7{,}000 meshes/sec at batch size 128.

\paragraph{Limitations}
Several limitations remain.
First, topology transfer quality depends on the quality of the source model's canonical mesh and the \name{} wrap registration: poorly registered wraps or source models with extreme vertex density asymmetry can degrade topology abstraction accuracy.
Second, despite pose-dependent correctives, standard LBS still produces artifacts at highly non-rigid deformations (\eg extreme elbow flexion, shoulder abduction beyond 90 degrees); learned correctives mitigate but do not eliminate these.
Third, adding a new identity backend requires implementing a new identity model class and a one-time \name{} mesh registration using standard non-rigid registration tools; this is a modest but non-trivial engineering step.
Fourth, \name{}'s pose abstraction is not a general-purpose retargeter: it recovers pose through mesh vertex correspondence, so it is limited to models that share compatible human body geometry.
It cannot abstract poses from characters with fundamentally different geometry or rigging structure (\eg robots, non-humanoid characters); such cases require a specialized retargeting solution.